\pdfoutput=1

\documentclass[10pt,letterpaper]{article}
\usepackage{amsmath,amssymb,bm}
\usepackage{booktabs,array,tabularx,multirow}
\usepackage{graphicx}
\usepackage{placeins}
\usepackage{fontawesome5}
\usepackage{styles/unimopd-jd}
\newcommand{\methodname}{Uni-MOPD}
\newcommand{\methodnameaccent}{\texorpdfstring{\textcolor{jdaccent}{\textsf{\bfseries Uni-MOPD}}}{Uni-MOPD}}
\usepackage{float}
\renewcommand{\topfraction}{0.9}

\title{\texorpdfstring{Strong Starts, Simple Recipes: \textcolor{jdaccent}{Uni}fied \textcolor{jdaccent}{M}ulti-Teacher\\[1mm]\textcolor{jdaccent}{O}n-\textcolor{jdaccent}{P}olicy \textcolor{jdaccent}{D}istillation from Multimodal Specialists}{Strong Starts, Simple Recipes: Unified Multi-Teacher On-Policy Distillation from Multimodal Specialists}}
\author[1,2]{Wenzheng~Chang}
\author[2,3]{Taiqiang~Wu}
\author[2]{Yuqi~Xu}
\author[2,*]{Yongfu~Zhu}
\author[2]{Zelin~Peng}
\author[2,4,6]{Ji~Zhao}
\author[2,5,6]{\\Shuai~Dong}
\author[2,7]{Wenpu~Liu}
\author[2,8]{Caoyuan~Ma}
\author[2]{Xuehui~Wang}
\author[2,7]{Ziyue~Wang}
\author[2,7]{Weichu~Xie}
\author[2,9]{\\Haiyuan~Wan}
\author[3]{Ngai~Wong}
\author[2]{Tong~He}
\author[2,\dagger]{Wenqi~Shao}
\author[2,*]{Jiaqi~Wang}
\affiliation[1]{Shanghai Jiao Tong University}
\affiliation[2]{JD.COM}
\affiliation[3]{The University of Hong Kong}
\affiliation[4]{University of Science and Technology of China}
\affiliation[5]{Fudan University}
\affiliation[6]{SII}
\affiliation[7]{Peking University}
\affiliation[8]{The University of Tokyo}
\affiliation[9]{Tsinghua University}
\contribution{$^{*}$Corresponding authors.\quad $^{\dagger}$Project Lead.}

\newcommand{\resourcelink}[3]{%
  \href{#1}{\color{jdaccent}%
    \tikz[baseline=-0.6ex]{\node[circle,draw=jdaccent,line width=0.5pt,
      inner sep=0pt,minimum size=18pt]{#2};}%
    \hspace{5pt}\underline{#3}}%
}
\newcommand{\projectlinks}{%
  \resourcelink{https://amberheart.github.io/Uni-MOPD/}{\faGlobe}{Project Page}%
  \hspace{20pt}%
  \resourcelink{https://github.com/AmberHeart/Uni-MOPD}{\faGithub}{GitHub}%
  \hspace{20pt}%
  \resourcelink{https://huggingface.co/collections/AmberHeart/uni-mopd}{\includegraphics[height=10pt]{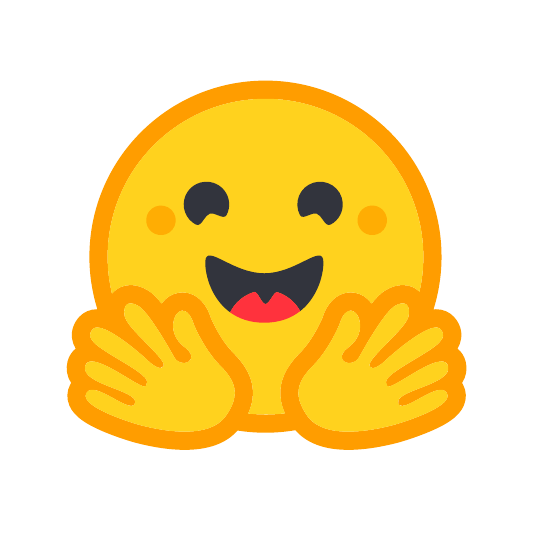}}{Hugging Face}%
  \hspace{20pt}%
  \resourcelink{https://modelscope.cn/collections/AmberHeart/Uni-MOPD}{\includegraphics[width=12pt]{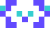}}{ModelScope}%
}
\hypersetup{pdftitle={Strong Starts, Simple Recipes: Unified Multi-Teacher On-Policy Distillation from Multimodal Specialists},pdfauthor={Wenzheng Chang, Taiqiang Wu, Yuqi Xu, Yongfu Zhu, Zelin Peng, Ji Zhao, Shuai Dong, Wenpu Liu, Caoyuan Ma, Xuehui Wang, Ziyue Wang, Weichu Xie, Haiyuan Wan, Ngai Wong, Tong He, Wenqi Shao, Jiaqi Wang}}
\abstract{Multi-teacher on-policy distillation (MOPD) integrates separately trained domain specialists into a single unified model. Existing methods focus on refining the distillation procedure while initializing the student from a base model or a multi-domain SFT checkpoint, which leaves capability integration largely to distillation itself. Instead, we treat the starting point as a design variable and propose Unified MOPD (\methodnameaccent{}), which pairs a strong start with a simple recipe: a training-free uniform merge of specialists from the same base model initializes the student, and short MOPD with uniform domain sampling and full-vocabulary reverse KL refines it. Across five multimodal domains and 19 benchmarks, our study establishes three findings. (1) The starting point matters more than subsequent training choices: before any training, the uniform merge already matches the score that the base model reaches after 200 MOPD updates, and after short distillation it matches higher-scoring, search-based starts. (2) Short MOPD recovers most of the capabilities the merge leaves unintegrated; it improves every domain, with the largest gains where the domain teacher still leads. (3) From a strong start, uniform domain sampling and full-vocabulary reverse KL match or exceed every evaluated sampling strategy and divergence, without schedules, per-token weighting, or truncation hyperparameters. \methodname{} outperforms Open-MOPD at the same training budget while training 3.3 times faster than its released code. We will release the training framework, specialist checkpoints, and unified student.\looseness=-1
}
\begin{document}
\maketitle

\section{Introduction}

Vision-language models must combine diverse capabilities, from mathematical reasoning and document understanding to GUI grounding and spatial reasoning. These capabilities often require different data and post-training procedures, so they are difficult to optimize jointly. A practical alternative trains domain specialists independently and then integrates them into a unified model. Multi-teacher on-policy distillation (MOPD) supports this route~\citep{Agarwal2024OPD,Ma2026MOPD}. The student generates responses to training queries, and the domain teacher of each query supervises them. MOPD has been adopted in large language model post-training~\citep{Xiao2026MiMoV2Flash,Xiaomi2026MiMoV26,DeepSeek2026V4,NVIDIA2026Nemotron3Ultra} and more recently in vision-language systems~\citep{KimiTeam2026K3,NVIDIA2026Nemotron35}.

Existing MOPD methods focus primarily on refining the distillation procedure, including domain sampling and scheduling~\citep{Gao2026OpenMOPD,Sun2026D3MOPD}, domain feedback normalization~\citep{Li2026DNMOPD}, and token-level teacher selection~\citep{Yin2026HOPD}. Many of these methods initialize the student from a base model~\citep{Li2026DNMOPD,Sun2026D3MOPD} or a multi-domain supervised fine-tuning (SFT) checkpoint~\citep{Gao2026OpenMOPD}. Initialization, however, matters in MOPD. It determines which capabilities the student already has, and it shapes the responses on which distillation takes place. A general checkpoint therefore leaves capability integration largely to subsequent training, even though the specialists already possess the target~capabilities.

Specialists trained from the same base model offer another possibility. Their parameters can be merged, without additional training, into a student that already combines capabilities across domains~\citep{Ilharco2023TaskArithmetic,Yu2024DARE}. This changes the role of subsequent distillation. Starting from a merged student, MOPD can focus on recovering the specialist capabilities that merging has not fully integrated. We study this route for multimodal specialists and treat the start as a design variable alongside the distillation recipe; text-only work instead uses merging as one component of new distillation algorithms~\citep{Kim2026IMMOPD,Yang2026CAMOPD}.

\begin{figure*}[!t]
  \centering
  \includegraphics[width=\linewidth]{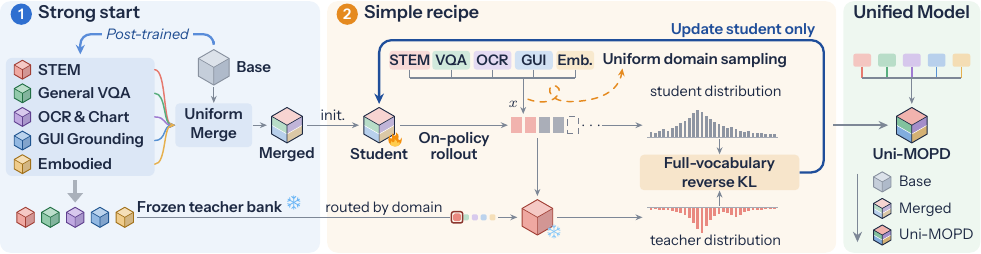}
  \caption{\textsf{\bfseries Pipeline of \methodnameaccent{}.} (1) Strong start: specialists post-trained from one base model are merged with equal weights, without training, and the merged checkpoint initializes the student. The same specialists then serve as the frozen teacher bank. (2) Simple recipe: the student generates on-policy responses to queries sampled uniformly across domains; the frozen domain teacher of each query scores the same responses, and full-vocabulary reverse KL updates only the student. Merging and short MOPD thus combine the five specialists into one unified model.}
  \label{fig:framework}
\end{figure*}

We propose \methodnameaccent{}, which pairs a strong start with a simple recipe (Figure~\ref{fig:framework}). As the strong start, a training-free \emph{uniform merge} combines the specialists with equal weights and no coefficient search. The simple recipe then refines this merged student through short on-policy distillation from the same specialists. It uses the standard reverse Kullback--Leibler (KL) divergence over the full vocabulary and \emph{uniform domain sampling}, which draws an equal number of training queries from every domain without a schedule.

Our study spans five multimodal domains: STEM reasoning, general visual question answering, document/OCR/chart understanding, GUI grounding, and embodied understanding. The specialist bank consists of five models post-trained from one instruction-tuned base model through distinct SFT and reinforcement learning (RL) paths. Each specialist leads its own domain, whereas a Multi-Domain SFT baseline trained on pooled data trails the corresponding specialist in every domain. Our goal is therefore to build a single model that approaches the strongest specialist everywhere. Evaluating starts, domain sampling, divergences, and full recipes on 19 benchmarks leads to three findings.

\textbf{(1) The starting point matters more than subsequent training choices.} Before any training, the uniform merge already matches the score that the base model reaches after 200 MOPD updates, and it exceeds Multi-Domain SFT. Its 9.5-point lead over the base model narrows only to 3.3 points after 50 updates, far exceeding the effect of any sampling or divergence choice we compare (Section~\ref{sec:query-selection}). Without coefficient search, the uniform merge matches the tuned merge, the best calibrated merges, and the strongest single specialist after short distillation.

\textbf{(2) Short distillation recovers the capabilities the start has not yet integrated.} From the uniform merge, short MOPD improves every domain, with the largest gains where the domain teacher still leads the merged student. The two stages thus play complementary roles: merging brings together capabilities across specialists, while distillation recovers most of the remaining teacher--student gap. When different specialists lead distinct facets of a domain, the two stages combine those complementary strengths rather than overwriting them.

\textbf{(3) A simple recipe suffices from a strong start.} Uniform domain sampling and full-vocabulary reverse KL match or exceed every evaluated sampling strategy and divergence. Among per-token objectives, distribution-level supervision outperforms sampled-token supervision, and full-vocabulary reverse KL supplies it exactly without a truncation hyperparameter. In 50 updates, \methodname{} closes 67\% of the gap from the uniform merge to the best-specialist reference, against 27\% for Open-MOPD~\citep{Gao2026OpenMOPD} under the same update and query budgets, whether compared against its released code or a controlled port on a shared backend. New strategies should therefore be evaluated from a strong start.

To make full-vocabulary MOPD practical at scale, we also develop a training framework for mixture-of-experts students that trains 3.3 times faster than the released Open-MOPD code on the same hardware (Section~\ref{sec:framework}). Across more than 150 experiments and 13,800 benchmark evaluations (over 60,000 NVIDIA H200 GPU-hours), these results establish a practical, search-free baseline for multimodal specialist integration. To support further work, we will release the training framework, specialist checkpoints, and unified student.

\section{Related Work}

\textbf{Multi-teacher on-policy distillation.} On-policy distillation
trains a student on its own outputs under teacher supervision~\citep{Agarwal2024OPD,gu2026minillmonpolicydistillationlarge}. MOPD
extends it to several teachers and routes each query to its domain
teacher~\citep{Ma2026MOPD},
whereas earlier multi-teacher methods distilled teacher distributions on
fixed data~\citep{Wan2024FuseLLM}.
Follow-up methods refine the procedure: Open-MOPD adds
token-share balancing, gap weighting, and reward refresh~\citep{Gao2026OpenMOPD}, D$^3$-MOPD schedules domain ratios
online~\citep{Sun2026D3MOPD}, H-OPD selects teachers per token rather
than per query~\citep{Yin2026HOPD}, and domain-normalized MOPD rescales
each domain's feedback~\citep{Li2026DNMOPD}. Others protect task-specific parameter
subspaces~\citep{Liu2026PMOPD}, use other teachers as
critics~\citep{Tian2026ROADMOPD}, distill each teacher's shift from its
own base~\citep{Xu2026DeltaMOPD}, or constrain the student toward a
moving average of itself~\citep{Yin2026SFMOPD}.

A second line concerns the divergence. Forward KL is commonly
described as mode-covering and reverse KL as mode-seeking~\citep{Wen2023fDistill,Ko2024DistiLLM}, though recent work
questions this view and studies how the two
differ in early optimization~\citep{Wu2025RethinkingKL}.
Single-teacher OPD methods adapt the divergence or target: EOPD adds
forward KL on high-entropy teacher tokens~\citep{Jin2026EOPD}, Veto
builds a geometric target in logit space~\citep{Jang2026Veto}, and
ToDi weights forward and reverse KL per
token~\citep{Jung2025ToDi}. Supervision granularity varies as well:
among large-scale MOPD systems, most supervise only
the sampled token~\citep{Xiao2026MiMoV2Flash,Yang2026NemotronCascade2,KimiTeam2026K3},
while DeepSeek-V4 distills over the full vocabulary~\citep{DeepSeek2026V4}. Because the divergence studies above are
single-teacher, we revisit these choices with multiple teachers and a
merged student. We compare three divergences (reverse KL, forward KL,
ToDi) and three supervision levels (sampled token, top-$k$, full
vocabulary).

The student's start is usually inherited from the surrounding
pipeline: the shared initial model~\citep{Li2026DNMOPD,Sun2026D3MOPD}, an SFT checkpoint~\citep{Ma2026MOPD,Gao2026OpenMOPD}, or a generalist RL checkpoint~\citep{NVIDIA2026Nemotron35}. When teachers come from separate
pipelines, a light SFT warmup precedes
distillation~\citep{NVIDIA2026Nemotron3Ultra}. Initialization has been
studied in adjacent stages, as teacher-compatible warm-up before
on-policy distillation~\citep{Liu2026SimpleOPD} or as preparation for
later reinforcement learning~\citep{Dong2026RLStartsBeforeRL}. In
contrast, we treat the start of MOPD as a design variable
and compare merged, single-specialist, and general starts under one
protocol.

\textbf{Model merging and MOPD.} Parameter merging combines
fine-tuned models that share a pre-trained base without additional training~\citep{Wortsman2022Soups,Ilharco2023TaskArithmetic}; Fisher merging
weights each parameter by its Fisher information~\citep{Matena2022Fisher}.
TIES reduces conflicts between task vectors by trimming and
sign election, whereas DARE randomly drops entries and rescales the rest~\citep{Yadav2023TIES,Yu2024DARE}. Merging also transfers skills across
vision-language models~\citep{Chen2025BringReason,Wei2026OptMerge}.
Data-dependent methods fit the coefficients on unlabeled data
through entropy minimization, activation alignment, or multi-teacher
distillation~\citep{Yang2024AdaMerging,Zhang2025ExpertMerging,Dalili2025SAMerging};
our offline calibration belongs to this family
(Section~\ref{sec:calibration-analysis}).

Merging has recently been combined with MOPD. Two text-only methods start
MOPD from a merge and add iterative task-vector updates~\citep{Kim2026IMMOPD} or capability-gap weighting~\citep{Yang2026CAMOPD}; from a strong start, the gap weighting we test
(Sections~\ref{sec:query-selection} and~\ref{sec:framework}) does not
improve on the simple recipe. Concurrent work learns the merge
coefficients during MOPD after finding that a uniform merge of text
specialists closes under half of the gap to the
teachers~\citep{Wang2026P2MOPD}; before any distillation, our multimodal merge closes 67\% of the Base-to-teacher gap under that per-domain measure (74\% in aggregate average score). A second-order analysis relates the MOPD
objective to Fisher-weighted merging~\citep{Liu2026FisherMOPD}, and other
text-only studies compare merging with mixed-data RL and distillation~\citep{Wu2026FusionParadigms,Wang2026MixMerge}. For text specialists sharing a base,
\citet{Xie2026RethinkMOPD} find that tuned off-policy distillation
matches MOPD and that weight merging recovers most of the gains.
For video specialists, MOPD outperforms parameter averaging~\citep{Qin2026VideoMOPD}. We show that the two stages are complementary for
multimodal specialists: the merge supplies the start, and MOPD recovers
what it leaves.

\textbf{Multimodal post-training and specialist integration.} Recent
language and vision-language models combine broad supervised training
with domain-specific reinforcement learning~\citep{Yang2025Qwen3,Bai2025Qwen3VL}. This domain-specific training produces separate
branches, and recent multimodal systems distill these branches on-policy
into a single student~\citep{KimiTeam2026K3,NVIDIA2026Nemotron35}. Our study isolates
this integration step with five specialists that follow
different post-training paths from a shared base checkpoint.

\section{Method}

Figure~\ref{fig:framework} outlines the two stages of \methodname{}. The
specialists are first merged in weight space to give the student a strong
start. The
specialists then serve as frozen domain teachers that supervise this
student during on-policy distillation. Section~\ref{sec:specialists}
describes the specialist bank, and Sections~\ref{sec:merge}
and~\ref{sec:mopd-queries} describe the two stages. For each stage, we
give the default choice and define the alternatives that
Section~\ref{sec:experiments} varies one at a time.

\subsection{Multimodal Specialist Models}\label{sec:specialists}

The specialist bank contains $K$ specialists $T_1,\dots,T_K$, each serving
as the teacher of one domain; $T_d$ denotes the teacher of domain $d$. We use $K=5$ domains: STEM reasoning,
general visual question answering, document/OCR/chart understanding, GUI
grounding, and embodied understanding, abbreviated in tables as STEM,
VQA, OCR, GUI, and Emb. All five specialists are post-trained from the
same base checkpoint (Base),
Qwen3-VL-30B-A3B-Instruct~\citep{Bai2025Qwen3VL}, an instruction-tuned
mixture-of-experts vision-language model. They follow different
post-training paths, each ending with reinforcement learning on its own
domain (Appendix~\ref{sec:specialist-bank}).

Because all specialists derive from one base model, they share an
architecture and vocabulary, which permits both weight-space merging and
token-level distillation. MOPD teachers are often obtained as domain RL
branches of one SFT checkpoint~\citep{Yang2026NemotronCascade2,Gao2026OpenMOPD}.
Our specialists instead follow distinct post-training paths and show
complementary strengths: each one leads its own domain and trails the
others elsewhere (Table~\ref{tab:main-results}). Specifically, the STEM,
VQA, and embodied specialists are trained with SFT and RL only on their
own domain, whereas the OCR and GUI specialists apply domain RL to
multi-domain SFT checkpoints, with the OCR specialist initialized from
the Multi-Domain SFT start of Section~\ref{sec:setup}, and are the
strongest overall.

\subsection{Strong Start from a Uniform Merge}\label{sec:merge}

Following task arithmetic~\citep{Ilharco2023TaskArithmetic}, we take
the shared base checkpoint $\theta_{\mathrm{base}}$ as the reference
point and define the task vector of specialist $k$ as
$\Delta_k=\theta_{T_k}-\theta_{\mathrm{base}}$. Each task vector is
sparsified and rescaled by DARE~\citep{Yu2024DARE} before summation:
\begin{equation}\label{eq:dare-merge}
\theta_0(\alpha)=\theta_{\mathrm{base}}+\sum_{k=1}^{K}\frac{\alpha_k}{1-\lambda}(m_k\odot\Delta_k),
\end{equation}
where $m_k\in\{0,1\}$ is a Bernoulli mask that keeps each entry with probability $1-\lambda$ ($\lambda$ is the drop rate), and $\alpha_k$ is the merge coefficient of
specialist $k$. The \textbf{uniform merge} gives every specialist the
same coefficient, $\alpha_k=1/K$ (0.2 for our five specialists), so the
total scale $\sum_k\alpha_k$ is one. It needs no specialist selection, no
coefficient search, and no training. The merges in our main comparisons
use a fixed drop rate and identical masks
(Appendix~\ref{sec:calibration-protocol}), so they differ only in
$\alpha$; Appendix~\ref{sec:coefficient-scale} examines sensitivity to the
drop rate and total scale.

Two alternatives replace the equal coefficients. The
\textbf{coefficient-tuned merge} (tuned merge) comes from a
manual search over two coefficient directions: the equal coefficients and
a hand-set direction that gives more weight to the two highest-scoring
specialists, OCR and GUI. Along each direction, we vary the total scale
and the DARE drop rate. The search is costly and draws on benchmark
evaluations. We keep the highest-scoring configuration within our manual
sweep on the full 19-benchmark suite before MOPD. The uniform
merge needs none of this. \textbf{Calibrated merges} fit
$\alpha$ offline to the teachers on teacher trajectories
(Section~\ref{sec:calibration-analysis}).

\subsection{Simple Recipe for On-Policy Distillation}\label{sec:mopd-queries}

The merged checkpoint $\theta_0$ initializes the student. Let
$p_\theta^h$ and $p_{T_k}^h$ denote the next-token distributions of the
student and of teacher $T_k$ over the shared vocabulary at prefix $h$.
For each query $x_i$ in a batch of $B$ queries, the student generates a
response $y_i$, and the domain teacher $T_{k(i)}$ scores the student's
prefixes $h_{it}=(x_i,y_{i,<t})$ at the valid response positions
$t\in\mathcal V_i$. With $N=\sum_{i=1}^B |\mathcal V_i|$ valid tokens in
the batch, we minimize the full-vocabulary reverse KL (RKL), averaged
over all valid response tokens in the batch:
\begin{equation}\label{eq:mopd-loss}
\mathcal L_{\mathrm{MOPD}}(\theta)
=\frac{1}{N}\sum_{i=1}^{B}\sum_{t\in\mathcal V_i}
D_{\mathrm{KL}}(p_\theta^{h_{it}}\Vert p_{T_{k(i)}}^{h_{it}}).
\end{equation}
Responses are held fixed within each update, and gradients flow only
through the student. Because the divergence is computed over the full
shared vocabulary, as in the large-scale MOPD of
DeepSeek-V4~\citep{DeepSeek2026V4}, every position is supervised by the
teacher's complete distribution rather than by a sampled-token or
top-$k$ estimate.
Section~\ref{sec:query-selection} compares this objective with forward KL
(FKL), ToDi~\citep{Jung2025ToDi}, and sampled-token and top-$k$
estimates.

Training queries come from the specialists' post-training data, and each
query is routed to the teacher of its source domain. The default
\textbf{uniform domain sampling} gives each of the $K$ domains the same
share $p_d=1/K$ of the training queries: it draws the same number of
queries from every domain, at random within each domain, and shuffles the
resulting pool globally. Uniform sampling thus balances query counts
across domains, while the $1/N$ normalization weights each domain by its
share of response tokens.

We compare uniform sampling with two families of non-uniform sampling
built from the same query pool. \textbf{Gap-weighted sampling} fixes the
domain shares before training according to the teacher--student gap
(the domain teacher's benchmark score minus that of the student's start
$\theta_0$). The mean of its positive parts over the benchmarks of domain $d$ is
$\eta_d=|\mathcal B_d|^{-1}\sum_{b\in\mathcal B_d}[s_b(T_d)-s_b(\theta_0)]_+$,
where $\mathcal B_d$ is the benchmark set of domain $d$ and $s_b$ the
score on benchmark $b$. The signals are fixed before training
(Appendix~\ref{sec:query-construction}). With $L_d$ the mean response length of domain
$d$,
\begin{equation}\label{eq:gap-sampling}
r_d=\eta_d/L_d^{\gamma},\qquad p_d=\frac{1}{2K}+\frac{r_d}{2\sum_j r_j},
\end{equation}
so half of the query budget is uniform and half follows the gap.
Gap-weighted sampling uses $\gamma=0$, and length-adjusted gap sampling
uses $\gamma=1/2$ to down-weight domains with long responses.

\textbf{Dynamic sampling} follows D$^3$-MOPD~\citep{Sun2026D3MOPD} and
updates the shares online from the training loss. For domain $d$, let
$\kappa_d$ be the smoothed MOPD loss, $\kappa_{d,0}$ its initial value,
and $\bar\rho_d$ its recent relative change. The scheduler favors domains
whose loss is still high and still falling. With $u_d$ the max-normalized
value of $(\kappa_d/\kappa_{d,0})\max(0,-\bar\rho_d)$,
\begin{equation}\label{eq:dynamic-sampling}
p_d=\frac{1}{2K}+\tfrac12\operatorname{softmax}_d(u/\tau),
\end{equation}
where $\tau$ is a temperature. Both families thus keep half of the
budget uniform, as the D$^3$-MOPD defaults do, so every domain stays
covered, and reallocate the other half by their own signal.

\textbf{Practical full-vocabulary distillation for MoE students.}
Evaluating Eq.~\eqref{eq:mopd-loss} over the full vocabulary for a
mixture-of-experts model is memory- and throughput-intensive when five
frozen teachers must score long multimodal responses. Our training
framework keeps the exact loss affordable through three design choices:
vocabulary logits are materialized only at response prediction positions
and the divergence is computed over vocabulary shards in chunks; frozen
teachers reside in CPU memory and are swapped onto the training GPUs one
domain at a time; and completed responses stream to teacher scoring while
the remaining rollouts finish (Appendix~\ref{sec:runtime-config},
Figure~\ref{fig:training-architecture}; throughput comparisons appear in
Section~\ref{sec:framework}).

\section{Experiments}\label{sec:experiments}

\begin{table}[!t]
\caption{\textsf{\bfseries Specialists and Starts.} Each specialist leads only its own domain, while the uniform merge matches the tuned merge after 50 MOPD updates. Top: the specialist bank and its best-specialist reference (the mean of the five bold entries). Bottom: each start before MOPD (Initial) and after 50 updates under the same protocol; Gain is the Avg.\ change from its own initial score. Shading ranks each column within its block (darker shading indicates a higher value) and bold marks the block maximum.}
\label{tab:main-results}
\centering\small
\setlength{\tabcolsep}{6pt}
\begin{tabular}{@{}lrrrrrrr@{}}
\toprule
 & STEM & VQA & OCR & GUI & Emb. & Avg. & Gain \\
\midrule
\multicolumn{8}{@{}l}{\textit{Specialist bank}} \\
\hspace{.5em}STEM & \tintcell{stem}{24}{\textbf{72.9}} & \tintcell{vqa}{19}{62.0} & \tintcell{ocr}{19}{77.5} & \tintcell{gui}{3}{15.9} & \tintcell{emb}{3}{51.5} & \tintcell{avg}{3}{56.0} & \\
\hspace{.5em}VQA & \tintcell{stem}{8}{63.1} & \tintcell{vqa}{24}{\textbf{66.4}} & \tintcell{ocr}{3}{56.6} & \tintcell{gui}{14}{60.4} & \tintcell{emb}{8}{52.4} & \tintcell{avg}{14}{59.8} & \\
\hspace{.5em}OCR & \tintcell{stem}{19}{71.5} & \tintcell{vqa}{14}{61.3} & \tintcell{ocr}{24}{\textbf{79.6}} & \tintcell{gui}{19}{76.1} & \tintcell{emb}{19}{53.3} & \tintcell{avg}{24}{\textbf{68.4}} & \\
\hspace{.5em}GUI & \tintcell{stem}{14}{67.7} & \tintcell{vqa}{8}{60.9} & \tintcell{ocr}{14}{76.3} & \tintcell{gui}{24}{\textbf{79.7}} & \tintcell{emb}{14}{53.1} & \tintcell{avg}{19}{67.5} & \\
\hspace{.5em}Embodied & \tintcell{stem}{3}{50.6} & \tintcell{vqa}{3}{58.4} & \tintcell{ocr}{8}{72.1} & \tintcell{gui}{8}{60.0} & \tintcell{emb}{24}{\textbf{56.0}} & \tintcell{avg}{8}{59.4} & \\
\hspace{.5em}\emph{Best-specialist ref.} & \emph{72.9} & \emph{66.4} & \emph{79.6} & \emph{79.7} & \emph{56.0} & \emph{70.9} & \\
\midrule
\multicolumn{8}{@{}l}{\textit{Base}} \\
\hspace{.5em}Initial & \tintcell{stem}{3}{58.3} & \tintcell{vqa}{3}{58.2} & \tintcell{ocr}{3}{71.8} & \tintcell{gui}{3}{51.3} & \tintcell{emb}{3}{51.0} & \tintcell{avg}{3}{58.1} &  \\
\hspace{.5em}MOPD @50 & \tintcell{stem}{6}{67.1} & \tintcell{vqa}{8}{61.5} & \tintcell{ocr}{5}{74.9} & \tintcell{gui}{6}{76.3} & \tintcell{emb}{6}{52.7} & \tintcell{avg}{7}{66.5} & +8.4 \\
\addlinespace[2pt]
\multicolumn{8}{@{}l}{\textit{Multi-Domain SFT}} \\
\hspace{.5em}Initial & \tintcell{stem}{5}{64.7} & \tintcell{vqa}{5}{59.7} & \tintcell{ocr}{8}{76.0} & \tintcell{gui}{10}{77.3} & \tintcell{emb}{5}{51.5} & \tintcell{avg}{5}{65.9} &  \\
\hspace{.5em}MOPD @50 & \tintcell{stem}{19}{70.6} & \tintcell{vqa}{15}{63.3} & \tintcell{ocr}{12}{77.6} & \tintcell{gui}{14}{78.1} & \tintcell{emb}{15}{55.5} & \tintcell{avg}{18}{69.0} & +3.2 \\
\addlinespace[2pt]
\multicolumn{8}{@{}l}{\textit{Tuned merge}} \\
\hspace{.5em}Initial & \tintcell{stem}{11}{68.7} & \tintcell{vqa}{12}{62.1} & \tintcell{ocr}{17}{77.9} & \tintcell{gui}{24}{\textbf{79.5}} & \tintcell{emb}{12}{54.4} & \tintcell{avg}{16}{68.5} &  \\
\hspace{.5em}MOPD @50 & \tintcell{stem}{22}{71.1} & \tintcell{vqa}{20}{64.8} & \tintcell{ocr}{17}{77.9} & \tintcell{gui}{22}{79.2} & \tintcell{emb}{19}{56.1} & \tintcell{avg}{24}{\textbf{69.8}} & +1.3 \\
\addlinespace[2pt]
\multicolumn{8}{@{}l}{\textit{OCR specialist}} \\
\hspace{.5em}Initial & \tintcell{stem}{24}{\textbf{71.5}} & \tintcell{vqa}{6}{61.3} & \tintcell{ocr}{24}{\textbf{79.6}} & \tintcell{gui}{5}{76.1} & \tintcell{emb}{8}{53.3} & \tintcell{avg}{14}{68.4} &  \\
\hspace{.5em}MOPD @50 & \tintcell{stem}{14}{69.9} & \tintcell{vqa}{24}{\textbf{65.3}} & \tintcell{ocr}{22}{78.8} & \tintcell{gui}{15}{78.2} & \tintcell{emb}{17}{56.0} & \tintcell{avg}{22}{69.7} & +1.3 \\
\addlinespace[2pt]
\multicolumn{8}{@{}l}{\textit{Uniform merge} (\methodnameaccent{})} \\
\hspace{.5em}Initial & \tintcell{stem}{8}{67.5} & \tintcell{vqa}{14}{62.3} & \tintcell{ocr}{10}{77.0} & \tintcell{gui}{8}{77.0} & \tintcell{emb}{10}{54.3} & \tintcell{avg}{9}{67.6} &  \\
\hspace{.5em}MOPD @50 & \tintcell{stem}{21}{71.0} & \tintcell{vqa}{22}{64.9} & \tintcell{ocr}{15}{77.8} & \tintcell{gui}{17}{78.3} & \tintcell{emb}{24}{\textbf{57.1}} & \tintcell{avg}{24}{\textbf{69.8}} & +2.2 \\
\bottomrule
\end{tabular}
\end{table}

The experiments follow the three findings of the Introduction: where to start, what short MOPD adds, and how to distill from a strong start. For the start, we compare student starts under the same MOPD protocol (Section~\ref{sec:main-results}) and show that calibrating the merge coefficients does not produce a better start (Section~\ref{sec:calibration-analysis}). For distillation, we measure what short MOPD adds to the uniform merge (Section~\ref{sec:what-mopd-adds}). For the recipe, we change one training choice at a time from the uniform merge (Section~\ref{sec:query-selection}) and compare the complete method and its training framework with Open-MOPD (Section~\ref{sec:framework}).

\subsection{Experimental Setup}
\label{sec:setup}

\textbf{Benchmarks and metric.} We evaluate on 19 public benchmarks
across the five domains of Section~\ref{sec:specialists}
(Appendix~\ref{sec:evaluation-suite}). All models use greedy decoding
with the same scoring and image preprocessing. Scores are on a 0--100 scale.
For a model $M$, we define the \textbf{domain-averaged score}
$S_{\mathrm{avg}}(M)$ by first averaging benchmarks within each domain
and then taking the unweighted mean of the five domain averages:
\[
S_d(M)=|\mathcal B_d|^{-1}\sum_{b\in\mathcal B_d}s_b(M),\qquad
S_{\mathrm{avg}}(M)=\tfrac{1}{K}\sum_{d=1}^{K}S_d(M).
\]
Avg.\ refers to $S_{\mathrm{avg}}$ throughout. Tables and figures show
one decimal, and differences and percentages are computed from unrounded scores.
Re-evaluating the same checkpoints changes Avg.\ by at most 0.1 and a
domain mean by at most 0.7, mainly from vLLM batch nondeterminism. Avg.\ differences above 0.15 and domain gaps above 0.7 therefore exceed
re-evaluation variation. Under domain- and source-balanced sampling, which
also balances data sources within each domain, two
uniform-merge runs with different seeds agree in Avg.\ to within 0.04,
well inside this margin, so each other training configuration is
reported at one seed.

The \textbf{per-domain best-specialist reference} (best-specialist reference for short) averages each domain's best specialist:
\[
S_{\mathrm{avg}}^\star=\tfrac{1}{K}\sum_d\max_{T\in\mathcal T}S_d(T)=70.9,
\]
where $\mathcal T$ is the specialist bank. It represents a hypothetical
unified model that matches the strongest specialist in every domain.

\textbf{Starts.} We compare five starts: Base without further training,
Multi-Domain SFT, the uniform and tuned merges of
Section~\ref{sec:merge}, and the OCR specialist (the strongest single
model in the bank). Multi-Domain SFT is trained from Base in a single
supervised run on pooled multi-domain data, without an RL stage
(Appendix~\ref{sec:specialist-bank}). Calibrated merges serve as further
starts in Section~\ref{sec:calibration-analysis}. To test robustness
without the two broad specialists (OCR and GUI),
Appendix~\ref{sec:four-teacher} also evaluates a second bank of four
narrow specialists trained strictly on their own domains (STEM, VQA,
OCR, and embodied) across its 16 benchmarks.

\textbf{Training.} Unless stated otherwise, runs use the following
protocol. Each optimizer update draws a fresh batch of 512 queries, so 50 updates
consume 25,600 queries and 200 consume 102,400. All comparisons report
update 50, a budget shared by all runs. All runs share
the same query sequence sampled uniformly across domains, the same teacher
routing, a fresh optimizer, a constant learning rate, and, unless a comparison
varies it, the RKL objective of Eq.~\eqref{eq:mopd-loss}. The student is trained with
ms-swift~\citep{Zhao2025SWIFT} on the Megatron-LM
backend~\citep{Shoeybi2019MegatronLM} with tensor and expert
parallelism. Rollouts come from a vLLM server~\citep{Kwon2023vLLM}, and
rollout routing replay (R3)~\citep{Ma2025R3} aligns mixture-of-experts routing
between generation and training (pipeline and comparison controls in
Appendices~\ref{sec:runtime-config} and~\ref{sec:comparison-controls}).

\textbf{Compute.} All experiments run on NVIDIA H200 GPUs. Every MOPD
run in this section, including the verl and Open-MOPD runs, uses three
nodes of eight GPUs (24 GPUs). In our framework, two nodes train the
student and one serves rollouts. We report cost in GPU-hours, the number
of allocated GPUs times the elapsed time, excluding checkpoint saving.
This counts the hardware a run occupies, not only its busy~time.

\subsection{Where to Start}
\label{sec:main-results}

\begin{figure*}[!t]\centering
\includegraphics[width=\textwidth]{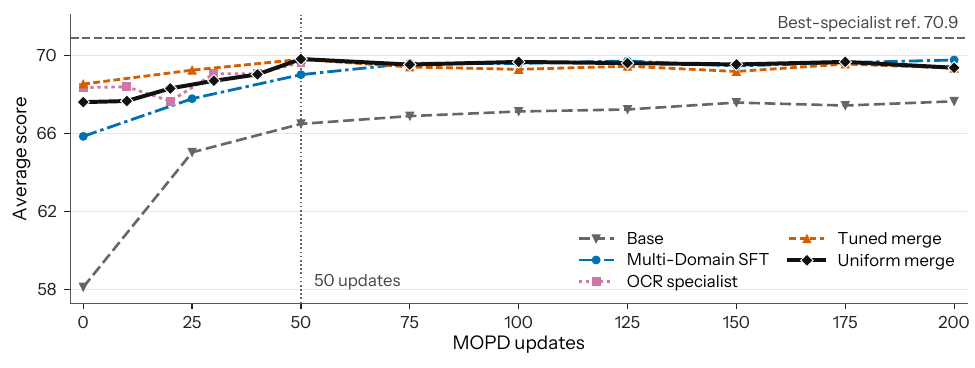}
\caption{\textsf{\bfseries Strong Starts under Short MOPD.} The uniform merge reaches the top level by update 50, whereas Base needs 200 updates to reach the initial score of the uniform merge. Average score of each start at every evaluated checkpoint; dashed: best-specialist reference; dotted: update 50. The OCR-specialist start is evaluated through update 50.}
\label{fig:merge-initialization}
\end{figure*}

We apply the same MOPD protocol to all five starts and report every
evaluated checkpoint
(Table~\ref{tab:main-results}, Figure~\ref{fig:merge-initialization}).
All starts except Base enter training well above it and count as
\emph{strong starts}.

\textbf{A training-free merge starts where Base ends.} The uniform
merge scores 67.6 before any training, 9.5 points above Base and 1.8
points above Multi-Domain SFT. It already matches what Base reaches
after 200 MOPD updates (67.7), and the merge closes 74\% of the gap
from Base to the best-specialist reference (70.9). It also stays within
5.4 points of the domain specialist in every domain
(Table~\ref{tab:main-results}). In contrast, the STEM, VQA, and embodied
specialists each fall more than 20 points below the best
specialist in at least one other domain.

\textbf{The uniform merge reaches the top level without search.} After
50 updates, the uniform merge leads Multi-Domain SFT by 0.8 points and
matches the tuned merge (69.8), which starts 0.9 points higher,
and the OCR-specialist start (69.7, from 68.4). Multi-Domain SFT needs
more than 100 updates to reach this level (Figure~\ref{fig:merge-initialization}).
Base does not reach this band: the Base-initialized student reaches 67.7
even after 200 updates, the score at which the merge starts.

\textbf{The merge rises steadily where the strongest specialist first
drops.} If a
strong start were enough on its own, one could simply begin from the
highest-scoring specialist, the OCR specialist in our bank. Its curve
first falls from 68.4 to 67.7 at update 20 as the STEM mean drops from
71.5 to 66.4, and it then recovers to 69.7 by update 50 (Figure~\ref{fig:merge-initialization}). The merge instead rises steadily, and its embodied mean
finishes 1.1 points above that of the OCR start. Under domain- and
source-balanced sampling, which spreads each domain's queries evenly
across its data sources, the uniform merge again finishes above the OCR
start (69.5 vs.\ 69.3; Appendix~\ref{sec:trajectory-details}). The
same pattern holds in the second bank of Appendix~\ref{sec:four-teacher},
whose four specialists are trained only on their own domains: there the
merge outperforms both Base and the strongest single specialist after 50
updates and leads the specialist start by 1.4 points at update 100.

\subsection{A Better-Fitting Merge Is Not a Better Start}
\label{sec:calibration-analysis}

Uniform weights give a strong start without any search, and optimizing
the coefficients does not improve on it after short MOPD.
Offline calibration fits the $K$ merge coefficients to the teachers, with
teacher weights, prefixes, and DARE masks fixed. For a prefix
$h=(x,z_{<t})$ of a teacher rollout $z$ and its domain teacher $T_k$, the
RKL objective over the rollout pool $\mathcal C$ is
\begin{equation}\label{eq:cal-loss}
\mathcal L_{\mathrm{cal}}(\alpha)
=\mathbb E_{(h,k)\sim\mathcal C}
D_{\mathrm{KL}}(p_{\theta_0(\alpha)}^h\Vert p_{T_k}^h).
\end{equation}
All three calibration families start from the uniform merge and reuse a
fixed pool of teacher rollouts (Appendix~\ref{sec:calibration-protocol}).
Projected gradient descent (PGD) minimizes the mean RKL of Eq.~\eqref{eq:cal-loss}. A maximin
quadratic program (QP) seeks a coefficient update that maximizes the
smallest relative RKL decrease across domains and accepts a step only if
every domain's RKL decreases. A four-objective study compares
RKL, cross-entropy (CE), FKL, and
ToDi~\citep{Jung2025ToDi}, and a separate Adam study calibrates the tuned
merge. Each family produces several checkpoints, so
Table~\ref{tab:merge-variants} keeps each family's best continued run
beside the uniform and tuned merges, and
Appendix~\ref{sec:calibration-details} gives the rest.

\begin{figure}[!t]\centering
\includegraphics[width=\textwidth]{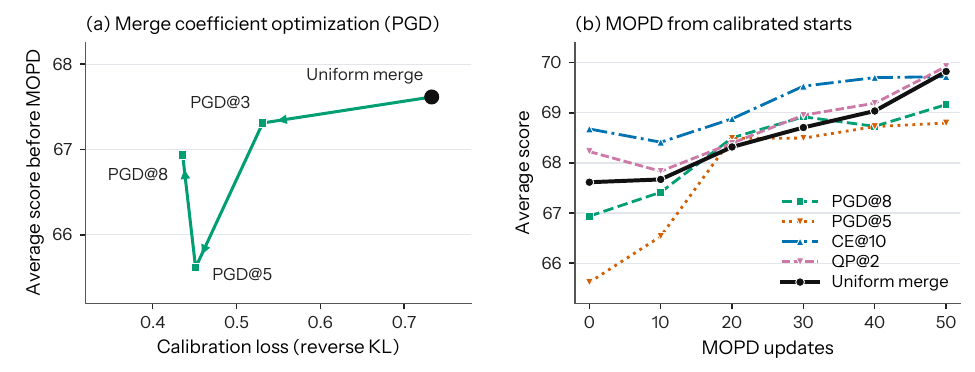}
\caption{\textsf{\bfseries Offline Calibration.} Offline calibration lowers the calibration loss without improving the 50-update outcome over the uniform merge. In (a), arrows follow the PGD coefficient updates away from the uniform merge; in (b), each calibrated merge is continued under the same MOPD recipe, with update 0 showing the score before MOPD.}
\label{fig:calibration-overview}
\end{figure}

\begin{table}[!t]
\caption{\textsf{\bfseries Calibrated Merges.} Offline-calibrated merges reach the same 50-update score as the uniform merge without coefficient search. Shown are Avg.\ before and after 50 MOPD updates for each calibration family's best continued start (suffix @$k$ counts coefficient updates) alongside the tuned and uniform merges; all nine calibrated merges appear in Appendix~\ref{sec:calibration-details}.}
\label{tab:merge-variants}
\centering\small
\setlength{\tabcolsep}{7pt}
\begin{tabular}{lrr}
\toprule
Merge & Initial & @50 \\
\midrule
Tuned & \tintcell{avg}{19}{68.5} & \tintcell{avg}{17}{69.8} \\
\midrule
PGD@8 & \tintcell{avg}{3}{66.9} & \tintcell{avg}{3}{69.2} \\
QP@2 & \tintcell{avg}{14}{68.2} & \tintcell{avg}{24}{\textbf{69.9}} \\
CE@10 & \tintcell{avg}{24}{\textbf{68.7}} & \tintcell{avg}{10}{69.7} \\
\midrule
Uniform merge & \tintcell{avg}{8}{67.6} & \tintcell{avg}{17}{69.8} \\
\bottomrule
\end{tabular}
\end{table}

\textbf{The uniform merge matches calibrated starts without coefficient
search.} Lower calibration loss does not give a better start. Along the
PGD path, the calibration RKL falls from 0.73 to
0.44, but the pre-MOPD Avg.\ moves to 67.3
(PGD@3), 65.6 (PGD@5), and 66.9 (PGD@8) and never exceeds the uniform
merge's 67.6 (Figure~\ref{fig:calibration-overview}a). The learned
coefficients show what the loss rewards. In the four-objective study,
10 RKL updates push the VQA coefficient down to 0.05 and the embodied
coefficient up to 0.41 and lower the pre-MOPD score by 0.7. In contrast,
CE@10 and FKL@7 keep their coefficients close to the tuned merge and
raise the pre-MOPD score by 1.1, but short MOPD removes that lead (69.7
and 69.6 vs.\ 69.8 for the uniform merge;
Appendix~\ref{sec:calibration-details}). Calibration
fits the merge on teacher-generated responses, whereas benchmarks score
the responses the model generates itself; we attribute the mismatch to
these different distributions.

\textbf{Short MOPD narrows the differences among merge starts.} CE@10
and QP@2 lose their initial lead and finish at the same level as the
uniform merge (Figure~\ref{fig:calibration-overview}b), while PGD@5 and
PGD@8, which start lower, recover most of the difference during MOPD.
Across all nine continued merges, the
3.1-point spread before MOPD shrinks to 1.1 after 50 updates. The
uniform merge matches the best calibrated start, QP@2, without any
search (Table~\ref{tab:merge-variants}). Neither manual nor loss-driven
coefficient search improves the 50-update outcome, so equal weighting
serves as a strong default.

\subsection{What Distillation Adds to a Strong Start}
\label{sec:what-mopd-adds}

Fifty MOPD updates take the uniform merge from 67.6 to 69.8 and raise
every domain mean (Table~\ref{tab:main-results}). Together, the merge and
the 50 updates close 92\% of the gap from Base to the best-specialist
reference, and the updates alone close 67\% of the remaining gap. On-policy
responses contribute 1.2 points of this gain: training on fixed teacher
responses reaches 68.6 (Appendix~\ref{sec:offpolicy}).

\begin{figure}[!t]\centering
\includegraphics[width=\textwidth]{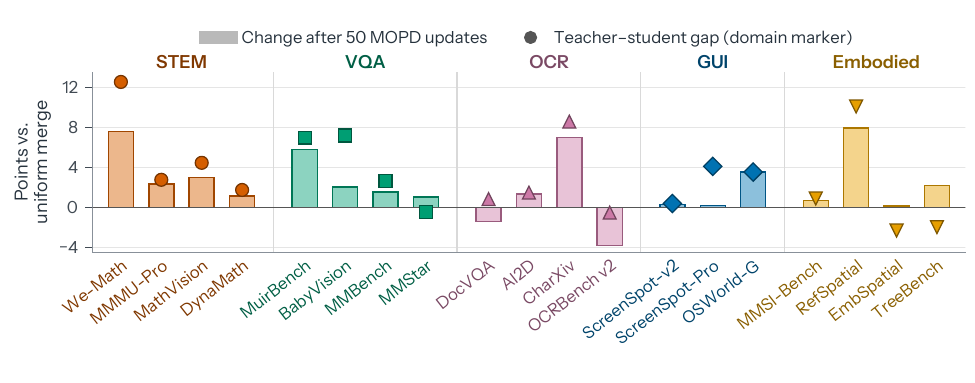}
\caption{\textsf{\bfseries Per-Benchmark Gains.} Gains are largest where the domain teacher leads the merge most. Bars show the score change on each benchmark after 50 MOPD updates from the uniform merge; markers show the teacher--student gap on the same benchmark, grouped by domain.}
\label{fig:post-merge}
\end{figure}

\textbf{Gains track the domain teacher's remaining lead.} Plotted
against the signed teacher--student gap on each benchmark
(Figure~\ref{fig:post-merge}), the gain after 50 updates has a Spearman
correlation of $\rho=0.77$ over the 19 benchmarks, computed from
unrounded scores. The largest gains occur where the teacher leads most:
We-Math gains 7.6 against a 12.6-point gap, RefSpatial gains 7.9 against
a 10.1-point gap, and CharXiv gains 7.0 against an 8.6-point gap. On
DocVQA and OCRBench v2, the merge already matches its teacher's level;
CharXiv and AI2D lift the OCR mean from 77.0 to 77.8.
TreeBench still gains 2.2 although its teacher scores
below the merge, consistent with the complementary strengths across the
bank discussed next.

\textbf{The unified model exceeds the embodied specialist.} The
embodied domain shows how the two stages combine. The embodied teacher
leads only on referring and multi-image spatial reasoning, with 70.4
on RefSpatial and 31.5 on MMSI-Bench. On TreeBench and
EmbSpatial-Bench, other specialists are stronger: the VQA specialist
scores 48.6 on TreeBench and the OCR specialist 80.6 on
EmbSpatial-Bench, against the embodied teacher's
44.2 and 77.9 (Appendix~\ref{sec:teacher-profiles}). The merge
retains those strengths from the bank, scoring 46.2 on TreeBench and
80.2 on EmbSpatial-Bench, above the embodied teacher on both. MOPD
keeps them, raises TreeBench to 48.4, and lifts RefSpatial
from 60.3 to 68.2, where that teacher has the largest remaining lead.
These changes raise the embodied mean
from 54.3 to 57.1, above the teacher's 56.0. The student stays at or
above the teacher from update 30 to 200
(Table~\ref{tab:main-run-trajectory}). The second bank, without the broad
OCR specialist, isolates this effect: there the embodied gain comes from
the embodied teacher alone (Appendix~\ref{sec:four-teacher}).

\textbf{Short budgets suffice.} The main run levels off by update 50 and
stays at 69.4--69.7 to update 200 as its loss keeps falling
(Table~\ref{tab:main-run-trajectory}). The
narrow bank does so by update 100 (Appendix~\ref{sec:four-teacher}).

\subsection{How to Distill from a Strong Start}
\label{sec:query-selection}

\begin{figure*}[!t]\centering
\includegraphics[width=\textwidth]{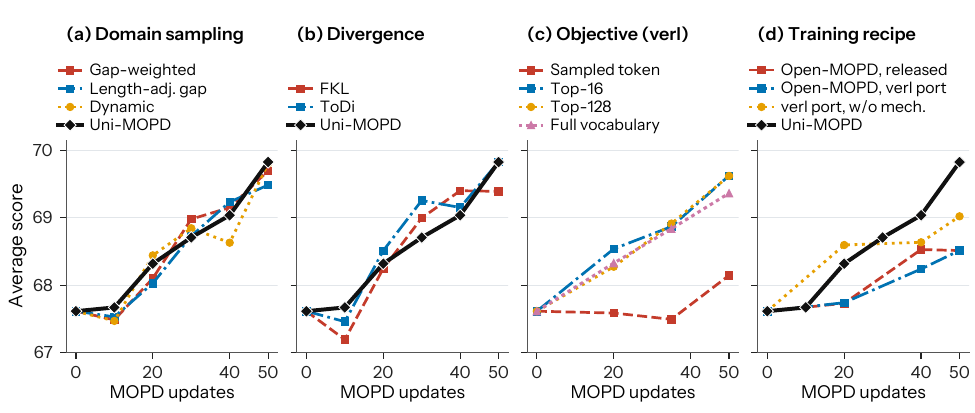}
\caption{\textsf{\bfseries Training Choices from a Strong Start.} Uniform sampling, reverse KL, and distribution-level supervision each match or outperform every alternative, and \methodname{} outperforms Open-MOPD. Panels (a)--(c) vary one training choice from the uniform merge; (d) compares \methodname{} with the released Open-MOPD code (FSDP2, no R3) and its \methodname{}-verl port (Megatron, R3), with and without the three budget-balancing mechanisms (update-50 scores in Tables~\ref{tab:fifty-update}--\ref{tab:framework}).}
\label{fig:training-choices}
\end{figure*}

Prior methods add sampling schedules, gap weighting, and
per-token divergence weighting. Each comparison fixes the start and
changes one factor: domain sampling, the divergence, or the per-token
supervision objective (Figure~\ref{fig:training-choices},
Table~\ref{tab:fifty-update}). None of the
added designs improves on the simple recipe. The recipe keeps only the elements that the ablations
show to contribute: on-policy responses, reverse KL, and distribution-level supervision
(Appendix~\ref{sec:offpolicy} and Table~\ref{tab:supervision}).

\textbf{Uniform sampling matches every reallocation without a
schedule.} Uniform and dynamic sampling both finish at 69.8 and
gap-weighted sampling at 69.7, while
length-adjusted gap sampling finishes 0.3 points lower at 69.5
(Figure~\ref{fig:training-choices}a). The strategies differ in
their domain profiles rather than their average. Uniform sampling keeps
every domain near 20\% of the queries. Gap-weighted sampling
shifts VQA and STEM to 23.7\% and 21.9\% and cuts OCR to 16.5\%,
and length-adjusted gap sampling raises GUI to 27.8\% while
STEM falls to 15.4\% (Appendix~\ref{sec:query-construction}).
Larger shares do not reliably raise the
scores: gap-weighted sampling lifts VQA (64.9 to 65.7) but lowers
STEM (71.0 to 70.0) despite its larger STEM share, whereas uniform
sampling keeps domains balanced at the top average. From Base,
gap-weighted and dynamic sampling reach 67.4 and 67.1 at update 200,
against 67.7 under uniform sampling (Table~\ref{tab:start-summary}).

\textbf{Reverse KL needs no per-token weighting.}
ToDi~\citep{Jung2025ToDi} interpolates between forward and reverse KL
with a per-token weight. Full-vocabulary RKL and ToDi both reach 69.8
at update 50, and FKL reaches 69.4 (Figure~\ref{fig:training-choices}b).
Standard reverse KL therefore matches ToDi with no per-token weighting
and leads FKL by 0.4 (Appendix~\ref{sec:runtime-todi}).

\textbf{Distribution-level supervision outperforms sampled-token
supervision.} To vary the per-token supervision objective alone, we
reimplement our method in verl~\citep{Sheng2025HybridFlow} (\methodname{}-verl) and compare a policy
gradient on the sampled token, the clipped policy-gradient surrogate of
Open-MOPD~\citep{Gao2026OpenMOPD} over the student's top 16 or top 128 tokens, and the exact full-vocabulary RKL.
In 50 updates, the sampled-token objective gains
0.5 points (Figure~\ref{fig:training-choices}c, Table~\ref{tab:supervision};
protocol in Appendix~\ref{sec:comparison-controls}), while the
top-16, top-128, and exact losses gain 1.7 to 2.0, more than three
times as much, and end within 0.3 points of one another (69.4--69.6).
Full-vocabulary RKL gives this supervision exactly at every position,
with no truncation size to tune. In our ms-swift framework, the same
full-vocabulary loss reaches 69.8, and Section~\ref{sec:framework}
confirms that our recipe also outperforms Open-MOPD within the shared
verl backend. Our framework runs this exact loss faster than
\methodname{}-verl runs top-16 (Table~\ref{tab:supervision-timing}).

\begin{table}[!t]
\begin{minipage}[t]{.495\linewidth}
\caption{\textsf{\bfseries Training Choices at 50 Updates.} Uniform sampling and RKL match or outperform every alternative from the uniform merge (Avg.\ 67.6). Each row varies one factor; shading and bold rank rows within each block as in Table~\ref{tab:main-results}.}
\label{tab:fifty-update}
\centering\footnotesize
\setlength{\tabcolsep}{2.5pt}
\begin{tabular}{lrrrrrr}
\toprule
Run & Avg. & STEM & VQA & OCR & GUI & Emb. \\
\midrule
\multicolumn{7}{@{}l}{\textit{Domain sampling}} \\
Uniform (\methodnameaccent{}) & \tintcell{avg}{24}{\textbf{69.8}} & \tintcell{stem}{24}{\textbf{71.0}} & \tintcell{vqa}{17}{64.9} & \tintcell{ocr}{24}{\textbf{77.8}} & \tintcell{gui}{3}{78.3} & \tintcell{emb}{17}{\textbf{57.1}} \\
Gap-weighted & \tintcell{avg}{10}{69.7} & \tintcell{stem}{3}{70.0} & \tintcell{vqa}{24}{\textbf{65.7}} & \tintcell{ocr}{10}{77.5} & \tintcell{gui}{17}{78.8} & \tintcell{emb}{10}{56.6} \\
Length-adjusted gap & \tintcell{avg}{3}{69.5} & \tintcell{stem}{17}{\textbf{71.0}} & \tintcell{vqa}{10}{64.5} & \tintcell{ocr}{3}{77.2} & \tintcell{gui}{10}{78.5} & \tintcell{emb}{3}{56.2} \\
Dynamic & \tintcell{avg}{17}{\textbf{69.8}} & \tintcell{stem}{10}{\textbf{71.0}} & \tintcell{vqa}{3}{64.1} & \tintcell{ocr}{17}{77.7} & \tintcell{gui}{24}{\textbf{79.1}} & \tintcell{emb}{24}{\textbf{57.1}} \\
\addlinespace[2pt]
\multicolumn{7}{@{}l}{\textit{Divergence}} \\
RKL (\methodnameaccent{}) & \tintcell{avg}{14}{\textbf{69.8}} & \tintcell{stem}{24}{\textbf{71.0}} & \tintcell{vqa}{3}{64.9} & \tintcell{ocr}{24}{\textbf{77.8}} & \tintcell{gui}{14}{\textbf{78.3}} & \tintcell{emb}{24}{\textbf{57.1}} \\
FKL & \tintcell{avg}{3}{69.4} & \tintcell{stem}{14}{70.9} & \tintcell{vqa}{14}{65.0} & \tintcell{ocr}{3}{77.2} & \tintcell{gui}{3}{\textbf{78.3}} & \tintcell{emb}{3}{55.6} \\
ToDi & \tintcell{avg}{24}{\textbf{69.8}} & \tintcell{stem}{3}{70.8} & \tintcell{vqa}{24}{\textbf{65.5}} & \tintcell{ocr}{14}{77.6} & \tintcell{gui}{24}{\textbf{78.3}} & \tintcell{emb}{14}{56.9} \\
\bottomrule
\end{tabular}
\end{minipage}\hfill
\begin{minipage}[t]{.465\linewidth}
\caption{\textsf{\bfseries Supervision Signals.} Distribution-level objectives outperform sampled-token supervision in every domain under \methodname{}-verl from the uniform merge (Avg.\ 67.6). Shading separates sampled-token from distribution-level supervision; the final \methodname{} row (same exact loss) is unranked.}
\label{tab:supervision}
\centering\footnotesize
\setlength{\tabcolsep}{3.0pt}
\begin{tabular}{lrrrrrr}
\toprule
Objective & Avg. & STEM & VQA & OCR & GUI & Emb. \\
\midrule
Sampled & \tintcell{avg}{3}{68.1} & \tintcell{stem}{3}{69.2} & \tintcell{vqa}{3}{62.3} & \tintcell{ocr}{3}{76.0} & \tintcell{gui}{3}{78.1} & \tintcell{emb}{3}{55.2} \\
Top-16 & \tintcell{avg}{24}{69.6} & \tintcell{stem}{24}{70.6} & \tintcell{vqa}{24}{64.9} & \tintcell{ocr}{24}{77.5} & \tintcell{gui}{24}{78.6} & \tintcell{emb}{24}{56.5} \\
Top-128 & \tintcell{avg}{24}{69.6} & \tintcell{stem}{24}{71.2} & \tintcell{vqa}{24}{64.9} & \tintcell{ocr}{24}{77.2} & \tintcell{gui}{24}{78.4} & \tintcell{emb}{24}{56.4} \\
Full vocab. & \tintcell{avg}{24}{69.4} & \tintcell{stem}{24}{70.3} & \tintcell{vqa}{24}{64.2} & \tintcell{ocr}{24}{77.8} & \tintcell{gui}{24}{78.5} & \tintcell{emb}{24}{56.0} \\
\midrule
\methodnameaccent{} & 69.8 & 71.0 & 64.9 & 77.8 & 78.3 & 57.1 \\
\bottomrule
\end{tabular}
\end{minipage}
\end{table}

\subsection{Comparison with Open-MOPD and Training Efficiency}
\label{sec:framework}

We compare the complete method with an existing MOPD recipe,
Open-MOPD~\citep{Gao2026OpenMOPD}. Open-MOPD distills through a clipped
policy-gradient loss over the student's top-16 tokens and adds three
budget-balancing mechanisms. Specifically, token-share balancing
equalizes each domain's share of response tokens in a batch; gap
weighting scales each domain by its mean per-token reward magnitude; and
reward refresh recomputes the student-dependent reward at each update
that reuses a rollout batch. We run the released code (verl, FSDP2), adapted to
multimodal MoE students and to the published recipe; it has no rollout
routing replay (R3; Section~\ref{sec:setup}). Because verl supports R3 only on
its Megatron and VeOmni backends, we also reimplement the recipe in
\methodname{}-verl (the verl port) and run it with and without the three
budget-balancing mechanisms. All runs start from the uniform merge
and train on our 25,600 queries in the same order at the same learning
rate. As in Open-MOPD, each rollout batch of 2,048 queries is split into
four updates of 512, so every run makes 50 updates of 512 queries
(Appendix~\ref{sec:comparison-controls}).

\begin{table}[!t]
\caption{\textsf{\bfseries Comparison with Open-MOPD.} \methodname{} outperforms every Open-MOPD variant while training 3.3$\times$ faster than the released code and 1.3$\times$ faster than the Megatron verl port. All runs complete 50 updates over the same 25,600 queries from the uniform merge on 24 H200 GPUs; w/o mech.\ removes the three budget-balancing mechanisms. The last two columns report seconds per update and H200 GPU-hours; shading and bold follow Table~\ref{tab:main-results}.}
\label{tab:framework}
\centering\small
\setlength{\tabcolsep}{5pt}
\begin{tabular}{@{}lrrrrrrrr@{}}
\toprule
Run & Avg. & STEM & VQA & OCR & GUI & Emb. & s/upd. & GPU-h \\
\midrule
Open-MOPD, released code & \tintcell{avg}{3}{68.5} & \tintcell{stem}{3}{67.4} & \tintcell{vqa}{3}{64.8} & \tintcell{ocr}{3}{76.2} & \tintcell{gui}{24}{\textbf{78.9}} & \tintcell{emb}{17}{55.3} & 888 & 296 \\
Open-MOPD, verl port & \tintcell{avg}{3}{68.5} & \tintcell{stem}{10}{67.8} & \tintcell{vqa}{24}{\textbf{65.4}} & \tintcell{ocr}{3}{76.2} & \tintcell{gui}{24}{\textbf{78.9}} & \tintcell{emb}{3}{54.2} & 348 & 116 \\
\quad w/o mech. & \tintcell{avg}{14}{69.0} & \tintcell{stem}{17}{69.3} & \tintcell{vqa}{14}{64.9} & \tintcell{ocr}{14}{76.9} & \tintcell{gui}{24}{\textbf{78.9}} & \tintcell{emb}{10}{55.0} & 351 & 117 \\
\midrule
\methodnameaccent{} & \tintcell{avg}{24}{\textbf{69.8}} & \tintcell{stem}{24}{\textbf{71.0}} & \tintcell{vqa}{14}{64.9} & \tintcell{ocr}{24}{\textbf{77.8}} & \tintcell{gui}{3}{78.3} & \tintcell{emb}{24}{\textbf{57.1}} & 266 & 89 \\
\bottomrule
\end{tabular}
\end{table}

\textbf{\methodname{} outperforms the Open-MOPD recipe with and without
routing replay.} Under matched update and query budgets, \methodname{}
reaches 69.8, while the released code and the verl port (with R3) both
reach 68.5 at update 50
(Table~\ref{tab:framework} and Figure~\ref{fig:training-choices}d).
On the same Megatron backend with R3, \methodname{}-verl with the same
full-vocabulary reverse KL reaches 69.4 (Table~\ref{tab:supervision},
Full vocab.\ row), 0.8 points above the verl port of Open-MOPD. This
difference comes from the recipe: one on-policy update per rollout batch
and no budget-balancing mechanisms.
\methodname{} closes 67\% of the 3.3-point
gap from the uniform merge (67.6) to the best-specialist
reference (70.9) in 50 updates, compared with 27\% for the Open-MOPD
recipe. The largest domain-level separation appears in STEM, where
\methodname{} reaches 71.0 while the two Open-MOPD implementations finish
at 67.4 and 67.8, close to the merge's 67.5 (domain and benchmark
breakdowns in Appendix~\ref{sec:openmopd-detail}). With 512-query rollout
batches, a 200-update run of the released code reaches 68.9, below our 50-update score. From the tuned
merge under its own default schedule, Open-MOPD without its mechanisms
needs 30,720 queries to reach the 69.7 score that \methodname{} reaches
from the same start with 25,600,
and the full Open-MOPD recipe never reaches it (peak 69.1;
Table~\ref{tab:openmopd-comparison}).

\textbf{From a strong start, Open-MOPD's budget-balancing mechanisms
reduce accuracy.} Removing
the three budget-balancing mechanisms raises the verl port at every
saved checkpoint (updates 20, 40, 50: 67.7$\to$68.6, 68.2$\to$68.6,
68.5$\to$69.0), with STEM accounting for most of the difference (69.3
vs.\ 67.8). Switching to one update per rollout batch with
token-averaged loss raises it further: with the port's own top-16
objective, the top-16 run of Table~\ref{tab:supervision} reaches 69.6
(Appendix~\ref{sec:comparison-controls}). Both the three mechanisms and
four updates per rollout batch with microbatch-averaged loss lower the
score from this start; the simple recipe keeps neither.

\textbf{Our framework reduces the cost of full-vocabulary distillation.}
Switching to the Megatron backend with routing replay and using our
streamed pipeline (Section~\ref{sec:mopd-queries} and
Appendix~\ref{sec:runtime-config}), our framework
takes 266 seconds per update and 89 GPU-hours for 50 updates, 3.3 times
faster than the released code (888 seconds and 296 GPU-hours), whose
FSDP2 actor update alone takes 628 seconds per update. In the verl port
(Megatron backend), the same Open-MOPD recipe takes 348 seconds and 116
GPU-hours, with an actor update of 210 seconds, and our framework is
1.3 times faster than this run as well. For the same full-vocabulary loss,
it is 2.3 times faster than \methodname{}-verl without CPU offloading
(over updates 2--15: 263 vs.\ 598 seconds per update;
Appendix~\ref{sec:comparison-controls}).

\section{Discussion}

\textbf{Where the gains come from.} The two elements the simple recipe
keeps, on-policy responses and distribution-level supervision, account for the
gains: removing either costs more than one point, more than any sampling
or divergence change. Because the start already supplies most of the
integration, new sampling or objective designs should be evaluated from
a strong start, so that their value is measured beyond what the start
provides.

\textbf{The start needs strength, not optimality.} Merges with higher
pre-distillation scores and the strongest single specialist all finish
within 0.2 points of the uniform merge after 50 updates
(Sections~\ref{sec:main-results} and~\ref{sec:calibration-analysis}).
Once weight merging places the student close to every domain teacher,
short on-policy distillation rapidly erases small initial coefficient
differences, so a training-free uniform merge serves as a strong default
without search or offline calibration.

\textbf{Why the two stages complement each other.} Merging combines the
specialists' updates relative to the shared base and retains strengths
across the entire bank. MOPD then has each domain teacher supervise the
merged student on its own responses, so updates concentrate where that
teacher still leads. When different specialists lead distinct benchmarks
within the same domain, merging preserves cross-specialist strengths and
distillation closes the remaining gaps to the domain teacher
(Section~\ref{sec:what-mopd-adds}).

\textbf{Why a simple recipe suffices.} Routing each query to its domain
teacher already directs supervision to the remaining gaps, and the gains
of short MOPD track these gaps (Figure~\ref{fig:post-merge}). Uniform
sampling then keeps balanced coverage across domains. Together these
choices match or exceed every evaluated reallocation and reweighting,
which move gains between domains without raising the average
(Section~\ref{sec:query-selection}).
Consistently, the Open-MOPD verl port reaches a stronger endpoint without its
budget-balancing mechanisms than with them
(Section~\ref{sec:framework}).

\textbf{Capability saturation precedes loss convergence.} In both banks,
benchmark scores saturate within a short budget, after 50 updates in the
main bank and around 100 in the second, while the distillation loss keeps
falling. This parallels the calibration result of
Section~\ref{sec:calibration-analysis}, where the calibration loss and
benchmark scores measure different response distributions: in both
stages, a lower distillation loss does not by itself mean higher
capability.

\textbf{Robustness.} The strong-start finding holds across two
specialist banks of different breadth post-trained from one
mixture-of-experts base model: five specialists evaluated on 19
benchmarks, and four narrow specialists on 16, where the merge leads the
single-specialist start by 1.4 points at update 100
(Appendix~\ref{sec:four-teacher}). The merge is also insensitive to its
sparsification setting: DARE drop rates from 0.3 to 0.9 change its
pre-MOPD score by at most 0.4 points
(Appendix~\ref{sec:coefficient-scale}). Uniform merging and uniform
domain sampling thus provide a search-free and schedule-free default
that matches benchmark-informed alternatives
(Appendices~\ref{sec:calibration-protocol}
and~\ref{sec:query-construction}).

\textbf{Limitations and scope.} Two boundaries govern our findings.
First, weight-space merging requires specialists that share a base
architecture, vocabulary, and base checkpoint, as in the domain RL
branches distilled by current MOPD systems; cross-architecture teacher
banks require an alternative initialization before distillation. Second,
all experiments study one 30B-A3B mixture-of-experts model family, and
most configurations are evaluated at a single random seed, grounded in
the 0.04-point seed spread measured under matched sampling
(Section~\ref{sec:setup}).

\FloatBarrier
\section{Conclusion}

When integrating multimodal specialists with MOPD, the starting point
has a larger effect than subsequent recipe choices. A training-free
uniform merge establishes a strong start that, after short
full-vocabulary MOPD, matches hand-tuned merges, offline-calibrated
merges, and the strongest single specialist without coefficient search.
Once distillation begins from this start, \methodname{} pairs uniform
domain sampling with full-vocabulary reverse KL to match or exceed every
evaluated sampling strategy and divergence. It outperforms Open-MOPD at
the same budget and closes 67\% of the remaining gap in 50 updates
compared with 27\% for Open-MOPD, while our framework trains 3.3 times
faster than the released Open-MOPD code.

The role of distillation depends on the start: it integrates the specialists'
capabilities from the base model and recovers what remains from a
strong one. Future MOPD methods should treat the start as a design
choice. To support further work, we will release the training framework,
which makes full-vocabulary MOPD efficient for mixture-of-experts
students, together with the specialist checkpoints and the unified
student.

\section*{LLM Usage Statement}
We used LLM-based tools (ChatGPT, Claude Code, and Codex) to assist with manuscript polishing, related-work and citation checking, and the development and debugging of experimental code. All AI-assisted content was reviewed and verified by the authors, who take full responsibility for the research findings and conclusions.

\bibliographystyle{plainnat}
{\renewcommand{\bibfont}{\fontsize{8.6}{10.2}\selectfont\raggedright}%
\setlength{\bibsep}{0.6pt plus .2pt minus .2pt}
\bibliography{references}

@inproceedings{Agarwal2024OPD,
  title = {On-Policy Distillation of Language Models: Learning from Self-Generated Mistakes},
  author = {Agarwal, Rishabh and Vieillard, Nino and Zhou, Yongchao and Stanczyk, Piotr and Ramos Garea, Sabela and Geist, Matthieu and Bachem, Olivier},
  booktitle = {International Conference on Learning Representations},
  year = {2024},
  eprint = {2306.13649},
  archivePrefix = {arXiv},
}

@inproceedings{Yu2024DARE,
  title = {Language Models are {S}uper {M}ario: Absorbing Abilities from Homologous Models as a Free Lunch},
  author = {Yu, Le and Yu, Bowen and Yu, Haiyang and Huang, Fei and Li, Yongbin},
  booktitle = {International Conference on Machine Learning},
  series = {Proceedings of Machine Learning Research},
  volume = {235},
  pages = {57755--57775},
  publisher = {PMLR},
  year = {2024},
  eprint = {2311.03099},
  archivePrefix = {arXiv},
}

@article{Kim2026IMMOPD,
  title = {No Pain, More Gain: Iterative Merging for Effective Multi-Teacher On-Policy Distillation},
  author = {Kim, Seonghyeon and Jang, Chaeyun and Lee, Noah and Kim, Boseop and Lee, Juho},
  journal = {arXiv preprint arXiv:2609.34745},
  year = {2026},
  eprint = {2609.34745},
  archivePrefix = {arXiv},
}

@inproceedings{Jung2025ToDi,
  title = {{ToDi}: Token-wise Distillation via Fine-Grained Divergence Control},
  author = {Jung, Seongryong and Yoon, Suwan and Kim, DongGeon and Lee, Hwanhee},
  booktitle = {Conference on Empirical Methods in Natural Language Processing},
  pages = {8078--8091},
  publisher = {Association for Computational Linguistics},
  year = {2025},
  doi = {10.18653/v1/2025.emnlp-main.409},
  eprint = {2505.16297},
  archivePrefix = {arXiv},
}

@article{Ma2026MOPD,
  title = {{MOPD}: Multi-Teacher On-Policy Distillation for Capability Integration in {LLM} Post-Training},
  author = {Ma, Wenhan and Wei, Jianyu and Zhao, Liang and Zhang, Hailin and Xiao, Bangjun and Li, Lei and Yang, Qibin and Gao, Bofei and Wang, Yudong and Li, Rang and others},
  journal = {arXiv preprint arXiv:2606.30406},
  year = {2026},
  eprint = {2606.30406},
  archivePrefix = {arXiv},
}

@article{Gao2026OpenMOPD,
  title = {{Open-MOPD}: Diagnosing and Fixing Capability Imbalance in Multi-Teacher On-Policy Distillation},
  author = {Gao, Huan-ang and Chi, Haohan and Yan, Yong and Feng, Shiyuan and Wu, Hanlin and Jiang, Zheng and He, Bingxiang and Ma, Wei-Ying and Zhang, Ya-Qin and Zhou, Hao},
  journal = {arXiv preprint arXiv:2608.19098},
  year = {2026},
  eprint = {2608.19098},
  archivePrefix = {arXiv},
}

@article{Bai2025Qwen3VL,
  title = {{Qwen3-VL} Technical Report},
  author = {Bai, Shuai and Cai, Yuxuan and Chen, Ruizhe and Chen, Keqin and Chen, Xionghui and Cheng, Zesen and Deng, Lianghao and Ding, Wei and Gao, Chang and Ge, Chunjiang and others},
  journal = {arXiv preprint arXiv:2511.21631},
  year = {2025},
  eprint = {2511.21631},
  archivePrefix = {arXiv},
}

@inproceedings{Sheng2025HybridFlow,
  title = {{HybridFlow}: A Flexible and Efficient {RLHF} Framework},
  author = {Sheng, Guangming and Zhang, Chi and Ye, Zilingfeng and Wu, Xibin and Zhang, Wang and Zhang, Ru and Peng, Yanghua and Lin, Haibin and Wu, Chuan},
  booktitle = {European Conference on Computer Systems},
  series = {EuroSys '25},
  pages = {1279--1297},
  publisher = {ACM},
  year = {2025},
  doi = {10.1145/3689031.3696075},
  eprint = {2409.19256},
  archivePrefix = {arXiv},
}

@inproceedings{Qiao2024WeMath,
  title = {{We-Math}: Does Your Large Multimodal Model Achieve Human-like Mathematical Reasoning?},
  author = {Qiao, Runqi and Tan, Qiuna and Dong, Guanting and Wu, Minhui and Sun, Chong and Song, Xiaoshuai and Wang, Jiapeng and GongQue, Zhuoma and Lei, Shanglin and Zhang, YiFan and others},
  booktitle = {Annual Meeting of the Association for Computational Linguistics},
  pages = {20023--20070},
  publisher = {Association for Computational Linguistics},
  year = {2025},
  doi = {10.18653/v1/2025.acl-long.983},
}

@inproceedings{Yue2024MMMUPro,
  title = {{MMMU-Pro}: A More Robust Multi-discipline Multimodal Understanding Benchmark},
  author = {Yue, Xiang and Zheng, Tianyu and Ni, Yuansheng and Wang, Yubo and Zhang, Kai and Tong, Shengbang and Sun, Yuxuan and Yu, Botao and Zhang, Ge and Sun, Huan and others},
  booktitle = {Annual Meeting of the Association for Computational Linguistics},
  pages = {15134--15186},
  publisher = {Association for Computational Linguistics},
  year = {2025},
  doi = {10.18653/v1/2025.acl-long.736},
}

@inproceedings{Wang2024MathVision,
  title = {Measuring Multimodal Mathematical Reasoning with {MATH-Vision} Dataset},
  author = {Wang, Ke and Pan, Junting and Shi, Weikang and Lu, Zimu and Ren, Houxing and Zhou, Aojun and Zhan, Mingjie and Li, Hongsheng},
  booktitle = {Advances in Neural Information Processing Systems},
  volume = {37},
  pages = {95095--95169},
  publisher = {Neural Information Processing Systems Foundation, Inc. (NeurIPS)},
  year = {2024},
  doi = {10.52202/079017-3014},
}

@inproceedings{Zou2024DynaMath,
  title = {{DynaMath}: A Dynamic Visual Benchmark for Evaluating Mathematical Reasoning Robustness of Vision Language Models},
  author = {Zou, Chengke and Guo, Xingang and Yang, Rui and Zhang, Junyu and Hu, Bin and Zhang, Huan},
  booktitle = {International Conference on Learning Representations},
  year = {2025},
}

@inproceedings{Wang2024MuirBench,
  title = {{MuirBench}: A Comprehensive Benchmark for Robust Multi-image Understanding},
  author = {Wang, Fei and Fu, Xingyu and Huang, James Y. and Li, Zekun and Liu, Qin and Liu, Xiaogeng and Ma, Mingyu Derek and Xu, Nan and Zhou, Wenxuan and Zhang, Kai and others},
  booktitle = {International Conference on Learning Representations},
  year = {2025},
}

@inproceedings{Chen2026BabyVision,
  title = {{BabyVision}: Visual Reasoning Beyond Language},
  author = {Chen, Liang and Xie, Weichu and Liang, Yiyan and He, Hongfeng and Zhao, Haozhe and Yang, Zhibo and Huang, Zhiqi and Wu, Haoning and Lu, Haoyu and Charles, Y. and others},
  booktitle = {International Conference on Machine Learning},
  series = {Proceedings of Machine Learning Research},
  volume = {306},
  pages = {17692--17713},
  publisher = {PMLR},
  year = {2026},
}

@inproceedings{Liu2023MMBench,
  title = {{MMBench}: Is Your Multi-Modal Model an All-around Player?},
  author = {Liu, Yuan and Duan, Haodong and Zhang, Yuanhan and Li, Bo and Zhang, Songyang and Zhao, Wangbo and Yuan, Yike and Wang, Jiaqi and He, Conghui and Liu, Ziwei and others},
  booktitle = {European Conference on Computer Vision},
  series = {Lecture Notes in Computer Science},
  volume = {15064},
  pages = {216--233},
  publisher = {Springer Nature Switzerland},
  year = {2025},
  doi = {10.1007/978-3-031-72658-3_13},
}

@inproceedings{Kembhavi2016AI2D,
  title = {A Diagram is Worth a Dozen Images},
  author = {Kembhavi, Aniruddha and Salvato, Mike and Kolve, Eric and Seo, Minjoon and Hajishirzi, Hannaneh and Farhadi, Ali},
  booktitle = {European Conference on Computer Vision},
  series = {Lecture Notes in Computer Science},
  volume = {9908},
  pages = {235--251},
  publisher = {Springer International Publishing},
  year = {2016},
  doi = {10.1007/978-3-319-46493-0_15},
}

@inproceedings{Mathew2020DocVQA,
  title = {{DocVQA}: A Dataset for {VQA} on Document Images},
  author = {Mathew, Minesh and Karatzas, Dimosthenis and Jawahar, C. V.},
  booktitle = {IEEE/CVF Winter Conference on Applications of Computer Vision},
  pages = {2200--2209},
  publisher = {IEEE},
  year = {2021},
  doi = {10.1109/WACV48630.2021.00225},
}

@inproceedings{Chen2024MMStar,
  title = {Are We on the Right Way for Evaluating Large Vision-Language Models?},
  author = {Chen, Lin and Li, Jinsong and Dong, Xiaoyi and Zhang, Pan and Zang, Yuhang and Chen, Zehui and Duan, Haodong and Wang, Jiaqi and Qiao, Yu and Lin, Dahua and others},
  booktitle = {Advances in Neural Information Processing Systems},
  volume = {37},
  pages = {27056--27087},
  publisher = {Neural Information Processing Systems Foundation, Inc. (NeurIPS)},
  year = {2024},
  doi = {10.52202/079017-0850},
}

@inproceedings{Wang2024CharXiv,
  title = {{CharXiv}: Charting Gaps in Realistic Chart Understanding in Multimodal {LLMs}},
  author = {Wang, Zirui and Xia, Mengzhou and He, Luxi and Chen, Howard and Liu, Yitao and Zhu, Richard and Liang, Kaiqu and Wu, Xindi and Liu, Haotian and Malladi, Sadhika and others},
  booktitle = {Advances in Neural Information Processing Systems},
  volume = {37},
  pages = {113569--113697},
  publisher = {Neural Information Processing Systems Foundation, Inc. (NeurIPS)},
  year = {2024},
  doi = {10.52202/079017-3609},
}

@inproceedings{Fu2025OCRBenchV2,
  title = {{OCRBench} v2: An Improved Benchmark for Evaluating Large Multimodal Models on Visual Text Localization and Reasoning},
  author = {Fu, Ling and Kuang, Zhebin and Song, Jiajun and Huang, Mingxin and Yang, Biao and Li, Yuzhe and Zhu, Linghao and Luo, Qidi and Wang, Xinyu and Lu, Hao and others},
  booktitle = {Advances in Neural Information Processing Systems},
  volume = {38},
  pages = {107930--107981},
  publisher = {Neural Information Processing Systems Foundation, Inc. (NeurIPS)},
  year = {2025},
  doi = {10.52202/085713-3253},
}

@inproceedings{Wu2024OSAtlas,
  title = {{OS-ATLAS}: Foundation Action Model for Generalist {GUI} Agents},
  author = {Wu, Zhiyong and Wu, Zhenyu and Xu, Fangzhi and Wang, Yian and Sun, Qiushi and Jia, Chengyou and Cheng, Kanzhi and Ding, Zichen and Chen, Liheng and Liang, Paul Pu and others},
  booktitle = {International Conference on Learning Representations},
  year = {2025},
}

@inproceedings{Li2025ScreenSpotPro,
  title = {{ScreenSpot-Pro}: {GUI} Grounding for Professional High-Resolution Computer Use},
  author = {Li, Kaixin and Meng, Ziyang and Lin, Hongzhan and Luo, Ziyang and Tian, Yuchen and Ma, Jing and Huang, Zhiyong and Chua, Tat-Seng},
  booktitle = {ACM Multimedia},
  pages = {8778--8786},
  publisher = {ACM},
  year = {2025},
  doi = {10.1145/3746027.3755688},
}

@inproceedings{Xie2025OSWorldG,
  title = {Scaling Computer-Use Grounding via User Interface Decomposition and Synthesis},
  author = {Xie, Tianbao and Deng, Jiaqi and Li, Xiaochuan and Yang, Junlin and Wu, Haoyuan and Chen, Jixuan and Hu, Wenjing and Wang, Xinyuan and Xu, Yuhui and Wang, Zekun and others},
  booktitle = {Advances in Neural Information Processing Systems},
  volume = {38},
  pages = {27405--27462},
  publisher = {Neural Information Processing Systems Foundation, Inc. (NeurIPS)},
  year = {2025},
  doi = {10.52202/085713-0817},
}

@inproceedings{Yang2025MMSI,
  title = {{MMSI-Bench}: A Benchmark for Multi-Image Spatial Intelligence},
  author = {Yang, Sihan and Xu, Runsen and Xie, Yiman and Yang, Sizhe and Li, Mo and Lin, Jingli and Zhu, Chenming and Chen, Xiaochen and Duan, Haodong and Yue, Xiangyu and others},
  booktitle = {International Conference on Learning Representations},
  year = {2026},
}

@inproceedings{Zhou2025RoboRefer,
  title = {{RoboRefer}: Towards Spatial Referring with Reasoning in Vision-Language Models for Robotics},
  author = {Zhou, Enshen and An, Jingkun and Chi, Cheng and Han, Yi and Rong, Shanyu and Zhang, Chi and Wang, Pengwei and Wang, Zhongyuan and Huang, Tiejun and Sheng, Lu and others},
  booktitle = {Advances in Neural Information Processing Systems},
  volume = {38},
  pages = {32209--32286},
  publisher = {Neural Information Processing Systems Foundation, Inc. (NeurIPS)},
  year = {2025},
  doi = {10.52202/085713-0957},
}

@inproceedings{Du2024EmbSpatial,
  title = {{EmbSpatial-Bench}: Benchmarking Spatial Understanding for Embodied Tasks with Large Vision-Language Models},
  author = {Du, Mengfei and Wu, Binhao and Li, Zejun and Huang, Xuanjing and Wei, Zhongyu},
  booktitle = {Annual Meeting of the Association for Computational Linguistics},
  pages = {346--355},
  publisher = {Association for Computational Linguistics},
  year = {2024},
  doi = {10.18653/v1/2024.acl-short.33},
}

@inproceedings{Wang2025TreeBench,
  title = {Traceable Evidence Enhanced Visual Grounded Reasoning: Evaluation and Method},
  author = {Wang, Haochen and Li, Xiangtai and Huang, Zilong and Wang, Anran and Wang, Jiacong and Zhang, Tao and Zheng, Jiani and Bai, Sule and Kang, Zijian and Feng, Jiashi and others},
  booktitle = {International Conference on Learning Representations},
  year = {2026},
}

@inproceedings{Ma2025R3,
  title = {Stabilizing {MoE} Reinforcement Learning by Aligning Training and Inference Routers},
  author = {Ma, Wenhan and Zhang, Hailin and Zhao, Liang and Song, Yifan and Wang, Yudong and Luo, Fuli and Sui, Zhifang},
  booktitle = {International Conference on Machine Learning},
  series = {Proceedings of Machine Learning Research},
  volume = {306},
  pages = {85320--85343},
  publisher = {PMLR},
  year = {2026},
}

@inproceedings{gu2026minillmonpolicydistillationlarge,
  title = {{MiniLLM}: Knowledge Distillation of Large Language Models},
  author = {Gu, Yuxian and Dong, Li and Wei, Furu and Huang, Minlie},
  booktitle = {International Conference on Learning Representations},
  year = {2024},
  eprint = {2306.08543},
  archivePrefix = {arXiv},
}

@inproceedings{Wan2024FuseLLM,
  title = {Knowledge Fusion of Large Language Models},
  author = {Wan, Fanqi and Huang, Xinting and Cai, Deng and Quan, Xiaojun and Bi, Wei and Shi, Shuming},
  booktitle = {International Conference on Learning Representations},
  year = {2024},
  eprint = {2401.10491},
  archivePrefix = {arXiv},
}

@inproceedings{Wen2023fDistill,
  title = {{$f$-Divergence} Minimization for Sequence-Level Knowledge Distillation},
  author = {Wen, Yuqiao and Li, Zichao and Du, Wenyu and Mou, Lili},
  booktitle = {Annual Meeting of the Association for Computational Linguistics},
  pages = {10817--10834},
  publisher = {Association for Computational Linguistics},
  year = {2023},
  doi = {10.18653/v1/2023.acl-long.605},
  eprint = {2307.15190},
  archivePrefix = {arXiv},
}

@inproceedings{Ko2024DistiLLM,
  title = {{DistiLLM}: Towards Streamlined Distillation for Large Language Models},
  author = {Ko, Jongwoo and Kim, Sungnyun and Chen, Tianyi and Yun, Se-Young},
  booktitle = {International Conference on Machine Learning},
  series = {Proceedings of Machine Learning Research},
  volume = {235},
  pages = {24872--24895},
  year = {2024},
  eprint = {2402.03898},
  archivePrefix = {arXiv},
}

@inproceedings{Wortsman2022Soups,
  title = {{Model Soups}: Averaging Weights of Multiple Fine-Tuned Models Improves Accuracy Without Increasing Inference Time},
  author = {Wortsman, Mitchell and Ilharco, Gabriel and Gadre, Samir Yitzhak and Roelofs, Rebecca and Gontijo-Lopes, Raphael and Morcos, Ari S and Namkoong, Hongseok and Farhadi, Ali and Carmon, Yair and Kornblith, Simon and Schmidt, Ludwig},
  booktitle = {International Conference on Machine Learning},
  series = {Proceedings of Machine Learning Research},
  volume = {162},
  pages = {23965--23998},
  year = {2022},
  eprint = {2203.05482},
  archivePrefix = {arXiv},
}

@inproceedings{Ilharco2023TaskArithmetic,
  title = {Editing Models with Task Arithmetic},
  author = {Ilharco, Gabriel and Ribeiro, Marco Tulio and Wortsman, Mitchell and Gururangan, Suchin and Schmidt, Ludwig and Hajishirzi, Hannaneh and Farhadi, Ali},
  booktitle = {International Conference on Learning Representations},
  year = {2023},
  eprint = {2212.04089},
  archivePrefix = {arXiv},
}

@inproceedings{Yadav2023TIES,
  title = {{TIES-Merging}: Resolving Interference When Merging Models},
  author = {Yadav, Prateek and Tam, Derek and Choshen, Leshem and Raffel, Colin and Bansal, Mohit},
  booktitle = {Advances in Neural Information Processing Systems},
  volume = {36},
  pages = {7093--7115},
  publisher = {Neural Information Processing Systems Foundation, Inc. (NeurIPS)},
  year = {2023},
  doi = {10.52202/075280-0310},
  eprint = {2306.01708},
  archivePrefix = {arXiv},
}

@inproceedings{Yang2024AdaMerging,
  title = {{AdaMerging}: Adaptive Model Merging for Multi-Task Learning},
  author = {Yang, Enneng and Wang, Zhenyi and Shen, Li and Liu, Shiwei and Guo, Guibing and Wang, Xingwei and Tao, Dacheng},
  booktitle = {International Conference on Learning Representations},
  year = {2024},
  eprint = {2310.02575},
  archivePrefix = {arXiv},
}

@inproceedings{Wei2026OptMerge,
  title = {{OptMerge}: Unifying Multimodal {LLM} Capabilities and Modalities via Model Merging},
  author = {Wei, Yongxian and Cheng, Runxi and Jin, Weike and Yang, Enneng and Shen, Li and Hou, Lu and Du, Sinan and Yuan, Chun and Cao, Xiaochun and Tao, Dacheng},
  booktitle = {International Conference on Learning Representations},
  year = {2026},
  eprint = {2505.19892},
  archivePrefix = {arXiv},
}

@article{Yang2025Qwen3,
  title = {{Qwen3} Technical Report},
  author = {Yang, An and Li, Anfeng and Yang, Baosong and Zhang, Beichen and Hui, Binyuan and Zheng, Bo and Yu, Bowen and Gao, Chang and Huang, Chengen and Lv, Chenxu and others},
  journal = {arXiv preprint arXiv:2505.09388},
  year = {2025},
  eprint = {2505.09388},
  archivePrefix = {arXiv},
}

@article{Wang2026MixMerge,
  title = {To Mix or To Merge: Toward Multi-Domain Reinforcement Learning for Large Language Models},
  author = {Wang, Haoqing and Long, Xiang and Li, Ziheng and Xu, Yilong and Li, Tingguang and Tang, Yehui},
  journal = {arXiv preprint arXiv:2602.12566},
  year = {2026},
  eprint = {2602.12566},
  archivePrefix = {arXiv},
}

@inproceedings{Yin2026HOPD,
  title = {{H-OPD}: Confidence Aware Heterogeneous Multi-Teacher Multimodal On-Policy Distillation},
  author = {Yin, Qixiang and Yao, Huanjin and Cai, Yuchen and Chen, Jianghao and Wang, Ziyi and Yang, Min and Su, Fei and Zhao, Zhicheng},
  booktitle = {Conference on Empirical Methods in Natural Language Processing},
  pages = {4780--4797},
  publisher = {Association for Computational Linguistics},
  year = {2026},
  eprint = {2607.02592},
  archivePrefix = {arXiv},
}

@inproceedings{Jin2026EOPD,
  title = {Entropy-Aware On-Policy Distillation of Language Models},
  author = {Jin, Woogyeol and Min, Taywon and Yang, Yongjin and Wei, Dennis and Zhou, Yi and Kadhe, Swanand Ravindra and Baracaldo, Nathalie and Lee, Kimin},
  booktitle = {International Conference on Machine Learning},
  series = {Proceedings of Machine Learning Research},
  volume = {306},
  pages = {53926--53943},
  publisher = {PMLR},
  year = {2026},
  eprint = {2603.07079},
  archivePrefix = {arXiv},
}

@inproceedings{Jang2026Veto,
  title = {Stable On-Policy Distillation through Adaptive Target Reformulation},
  author = {Jang, Ijun and Yeom, Jewon and Yeo, Juan and Lim, Hyunggu and Kim, Taesup},
  booktitle = {Findings of the Association for Computational Linguistics: ACL 2026},
  pages = {42217--42227},
  publisher = {Association for Computational Linguistics},
  year = {2026},
  doi = {10.18653/v1/2026.findings-acl.2094},
}

@inproceedings{Chen2025BringReason,
  title = {Bring Reason to Vision: Understanding Perception and Reasoning through Model Merging},
  author = {Chen, Shiqi and Zhang, Jinghan and Zhu, Tongyao and Liu, Wei and Gao, Siyang and Xiong, Miao and Li, Manling and He, Junxian},
  booktitle = {International Conference on Machine Learning},
  series = {Proceedings of Machine Learning Research},
  volume = {267},
  pages = {9803--9817},
  year = {2025},
  eprint = {2505.05464},
  archivePrefix = {arXiv},
}

@inproceedings{Zhang2025ExpertMerging,
  title = {Expert Merging: Model Merging with Unsupervised Expert Alignment and Importance-Guided Layer Chunking},
  author = {Zhang, Dengming and Ma, Xiaowen and Ni, Zhenliang and Wu, Zhenkai and Shu, Han and Jiang, Xin and Chen, Xinghao},
  booktitle = {International Conference on Learning Representations},
  year = {2026},
}

@article{Zhao2025SWIFT,
  title = {{SWIFT}: A Scalable Lightweight Infrastructure for Fine-Tuning},
  author = {Zhao, Yuze and Huang, Jintao and Hu, Jinghan and Wang, Xingjun and Mao, Yunlin and Zhang, Daoze and Jiang, Zeyinzi and Wu, Zhikai and Ai, Baole and Wang, Ang and others},
  journal = {Proceedings of the AAAI Conference on Artificial Intelligence},
  volume = {39},
  number = {28},
  pages = {29733--29735},
  publisher = {Association for the Advancement of Artificial Intelligence (AAAI)},
  year = {2025},
  doi = {10.1609/aaai.v39i28.35383},
}

@article{Shoeybi2019MegatronLM,
  title = {{Megatron-LM}: Training Multi-Billion Parameter Language Models Using Model Parallelism},
  author = {Shoeybi, Mohammad and Patwary, Mostofa and Puri, Raul and LeGresley, Patrick and Casper, Jared and Catanzaro, Bryan},
  journal = {arXiv preprint arXiv:1909.08053},
  year = {2019},
  eprint = {1909.08053},
  archivePrefix = {arXiv},
}

@inproceedings{Kwon2023vLLM,
  title = {Efficient Memory Management for Large Language Model Serving with {PagedAttention}},
  author = {Kwon, Woosuk and Li, Zhuohan and Zhuang, Siyuan and Sheng, Ying and Zheng, Lianmin and Yu, Cody Hao and Gonzalez, Joseph E. and Zhang, Hao and Stoica, Ion},
  booktitle = {Symposium on Operating Systems Principles},
  pages = {611--626},
  publisher = {ACM},
  year = {2023},
  doi = {10.1145/3600006.3613165},
}

@article{Wu2026FusionParadigms,
  title = {Consolidating {RLVR} Capabilities Across Domains: A Deep Dive into Fusion Paradigms},
  author = {Wu, Siye and Yang, Kai and Cai, Yuchen and Xu, Xin and Wang, Peng-Yuan and Wang, Jiaxuan and Liu, Jiashun and Lyu, Jiafei and Chen, Yangkun and Yang, Saiyong and others},
  journal = {arXiv preprint arXiv:2608.27409},
  year = {2026},
  eprint = {2608.27409},
  archivePrefix = {arXiv},
}

@article{Liu2026FisherMOPD,
  title = {A {Fisher}-Geometric View of Multi-Teacher On-Policy Distillation},
  author = {Liu, Kaiyuan and Zhuang, Ziyuan and Ma, Haoxuan and Gu, Yannian and Song, Mingfei and Weng, Rongxiang and Ye, Jieping},
  journal = {arXiv preprint arXiv:2610.26525},
  year = {2026},
  eprint = {2610.26525},
  archivePrefix = {arXiv},
}

@article{Xiao2026MiMoV2Flash,
  title = {{MiMo-V2-Flash} Technical Report},
  author = {{Xiaomi LLM-Core Team}},
  journal = {arXiv preprint arXiv:2601.02780},
  year = {2026},
  eprint = {2601.02780},
  archivePrefix = {arXiv},
}

@article{Xiaomi2026MiMoV26,
  title = {{MiMo-V2.6}: Scaling Reinforcement Learning Towards Self-Improvement},
  author = {{Xiaomi LLM-Core Team}},
  journal = {arXiv preprint arXiv:2610.11959},
  year = {2026},
  eprint = {2610.11959},
  archivePrefix = {arXiv},
}

@article{Yang2026NemotronCascade2,
  title = {{Nemotron-Cascade 2}: Post-Training {LLMs} with Cascade {RL} and Multi-Domain On-Policy Distillation},
  author = {Yang, Zhuolin and Liu, Zihan and Chen, Yang and Dai, Wenliang and Wang, Boxin and Lin, Sheng-Chieh and Lee, Chankyu and Chen, Yangyi and Jiang, Dongfu and He, Jiafan and others},
  journal = {arXiv preprint arXiv:2603.19220},
  year = {2026},
  eprint = {2603.19220},
  archivePrefix = {arXiv},
}

@article{NVIDIA2026Nemotron3Ultra,
  title = {{Nemotron 3 Ultra}: Open, Efficient Mixture-of-Experts Hybrid {Mamba-Transformer} Model for Agentic Reasoning},
  author = {{NVIDIA} and Blakeman, Aaron and Thomas, Aaron and Jhunjhunwala, Aastha and others},
  journal = {arXiv preprint arXiv:2606.15007},
  year = {2026},
  eprint = {2606.15007},
  archivePrefix = {arXiv},
}

@article{Li2026DNMOPD,
  title = {Beyond Teacher Assignment: Domain-Normalized Multi-Teacher On-Policy Distillation},
  author = {Li, Xin and Jiang, Hao and Gao, Xin and Wang, Annan and Xie, Yuchen and Guo, Jinghao and Qu, Xingwei and Zhang, Yichi and Yuen, Chau},
  journal = {arXiv preprint arXiv:2609.35347},
  year = {2026},
  eprint = {2609.35347},
  archivePrefix = {arXiv},
}

@inproceedings{Matena2022Fisher,
  title = {Merging Models with {Fisher}-Weighted Averaging},
  author = {Matena, Michael S. and Raffel, Colin},
  booktitle = {Advances in Neural Information Processing Systems},
  volume = {35},
  pages = {17703--17716},
  publisher = {Neural Information Processing Systems Foundation, Inc. (NeurIPS)},
  year = {2022},
  doi = {10.52202/068431-1287},
  eprint = {2111.09832},
  archivePrefix = {arXiv},
}

@article{Sun2026D3MOPD,
  title = {{D$^3$-MOPD}: Dynamic Domain {ScheDuling} for Efficient Multi-Teacher Distillation},
  author = {{AllSpark Team}},
  journal = {arXiv preprint arXiv:2608.24987},
  year = {2026},
  eprint = {2608.24987},
  archivePrefix = {arXiv},
}

@article{Yang2026CAMOPD,
  title = {Beyond Policy Discrepancy: Capability-Aware Multi-Teacher On-Policy Distillation},
  author = {Yang, Xuanfeng and Gai, Le and Wang, Jiaming and Zhang, Xuan},
  journal = {arXiv preprint arXiv:2610.26950},
  year = {2026},
  eprint = {2610.26950},
  archivePrefix = {arXiv},
}

@article{Dalili2025SAMerging,
  title = {Model Merging via Multi-Teacher Knowledge Distillation},
  author = {Dalili, Seyed Arshan and Mahdavi, Mehrdad},
  journal = {arXiv preprint arXiv:2512.21288},
  year = {2025},
  eprint = {2512.21288},
  archivePrefix = {arXiv},
}

@article{DeepSeek2026V4,
  title = {{DeepSeek-V4}: Towards Highly Efficient Million-Token Context Intelligence},
  author = {{DeepSeek-AI} and Xu, Anyi and Lin, Bangcai and Xue, Bing and Wang, Bingxuan and others},
  journal = {arXiv preprint arXiv:2606.19348},
  year = {2026},
  eprint = {2606.19348},
  archivePrefix = {arXiv},
}

@article{KimiTeam2026K3,
  title = {{Kimi K3}: Open Frontier Intelligence},
  author = {{Kimi Team} and Bai, Tongtong and Bai, Yifan and Bao, Yiping and others},
  journal = {arXiv preprint arXiv:2607.24653},
  year = {2026},
  eprint = {2607.24653},
  archivePrefix = {arXiv},
}

@article{NVIDIA2026Nemotron35,
  title = {{Nemotron 3.5 Super} and {Lightning}: Open, Efficient Mixture-of-Experts Hybrid {Mamba-Transformer} Models for Agentic {AI}},
  author = {{NVIDIA}},
  journal = {arXiv preprint arXiv:2610.38095},
  year = {2026},
  eprint = {2610.38095},
  archivePrefix = {arXiv},
}

@inproceedings{Wu2025RethinkingKL,
  title = {Rethinking {Kullback-Leibler} Divergence in Knowledge Distillation for Large Language Models},
  author = {Wu, Taiqiang and Tao, Chaofan and Wang, Jiahao and Yang, Runming and Zhao, Zhe and Wong, Ngai},
  booktitle = {International Conference on Computational Linguistics},
  pages = {5737--5755},
  publisher = {Association for Computational Linguistics},
  address = {Abu Dhabi, UAE},
  year = {2025},
}

@article{Liu2026SimpleOPD,
  title = {{Simple-OPD}: Demystifying Warm-up for On-policy Distillation},
  author = {Liu, Tao and Wu, Taiqiang and Zheng, Mao and Luo, Xuan and Yang, Runming and Yang, Xuewei and Wang, Junjie and Yang, Yujiu},
  journal = {arXiv preprint arXiv:2608.06802},
  year = {2026},
  eprint = {2608.06802},
  archivePrefix = {arXiv},
}

@article{Dong2026RLStartsBeforeRL,
  title = {{RL} Starts before {RL}: On Policy Distillation for Better Reinforcement Learning},
  author = {Dong, Shuai and Zhu, Yongfu and Xu, Yuqi and Xie, Weichu and Liuwenpu and Wang, Ziyue and Tuo, Kaiwen and Wang, Congcong and Wang, Siyuan and Shao, Wenqi and Yang, Shuai and others},
  journal = {arXiv preprint arXiv:2609.28145},
  year = {2026},
  eprint = {2609.28145},
  archivePrefix = {arXiv},
}

@article{Xie2026RethinkMOPD,
  title = {Rethinking Self-Distillation for Multi-Teacher Capability Merging},
  author = {Xie, Roy and Friedman, Dan and Nan, Feng and Huang, Yukun and Xu, Zhichao and Zhang, Chengjiu and Xu, Jun and Faruqui, Manaal and Rathod, Vivek and Dhingra, Bhuwan},
  journal = {arXiv preprint arXiv:2610.04272},
  year = {2026},
  eprint = {2610.04272},
  archivePrefix = {arXiv},
}

@article{Qin2026VideoMOPD,
  title = {{Video-MOPD}: Multi-Teacher On-Policy Distillation for Video Understanding},
  author = {Qin, Zhenxin and Shi, Peng and Han, Cong and Qian, Yinlong and Jie, Zequn and Ma, Lin},
  journal = {arXiv preprint arXiv:2609.09300},
  year = {2026},
  eprint = {2609.09300},
  archivePrefix = {arXiv},
}

@article{Yin2026SFMOPD,
  title = {Slow-Fast Multi-Teacher On-Policy Distillation for Capability Preservation},
  author = {Yin, Xiaofei and Chu, Tong and Fu, Jiyuan and Lan, Jun and Zhou, Shuheng and Zhu, Huijia},
  journal = {arXiv preprint arXiv:2610.02324},
  year = {2026},
  eprint = {2610.02324},
  archivePrefix = {arXiv},
}

@article{Xu2026DeltaMOPD,
  title = {Composing What Each Teacher Learned: Delta Distillation for Multi-Teacher On-Policy Learning},
  author = {Xu, Yuanda and Sang, Hejian and Zhou, Zhengze and Hamidi, Shayan Mohajer and Li, Xiaomin and Jain, Rohit and Geramifard, Alborz},
  journal = {arXiv preprint arXiv:2610.10460},
  year = {2026},
  eprint = {2610.10460},
  archivePrefix = {arXiv},
}

@article{Liu2026PMOPD,
  title = {{PMOPD}: Task Ordering, Cycling, and Parameter-Update Subspace Protection in Multi-Teacher On-Policy Distillation},
  author = {Liu, Youzhi and Zheng, Ruobing and Tong, Boyuan and Li, Tianqi and Li, Pingqi and Bi, Hanbo and Yuan, Yi and Chen, Jingdong},
  journal = {arXiv preprint arXiv:2609.34605},
  year = {2026},
  eprint = {2609.34605},
  archivePrefix = {arXiv},
}

@article{Tian2026ROADMOPD,
  title = {Beyond Domain-Routed Supervision: Cross-Teacher Guarding for Multi-Teacher On-Policy Distillation},
  author = {Tian, Shi-Yu and Liu, Yong and Liu, Ximan and Zhu, Tianrui and Bai, Sule and Yang, Xinwei and Liu, Ruiqi and Yu, Kun-Yang and Yang, Ming and Zhou, Zhi and others},
  journal = {arXiv preprint arXiv:2610.13521},
  year = {2026},
  eprint = {2610.13521},
  archivePrefix = {arXiv},
}

@article{Wang2026P2MOPD,
  title = {{P$^2$-MOPD}: Accelerating Multi-Teacher On-Policy Distillation via Parameter Priors},
  author = {Wang, Sinuo and Lu, Jiacheng and Zhang, Zeyu and Song, Maojia and Sun, Rui and Luan, Beidi and Xie, Yutong and Li, Jing and Jiang, Daxin and Wu, Qi and Bai, Zuo},
  journal = {arXiv preprint arXiv:2611.03953},
  year = {2026},
  eprint = {2611.03953},
  archivePrefix = {arXiv},
}
}

\onecolumn
\appendix
\setcounter{topnumber}{3}\setcounter{totalnumber}{4}\renewcommand{\topfraction}{0.85}
\pretocmd{\section}{\FloatBarrier}{}{}
\numberwithin{table}{section}
\numberwithin{figure}{section}
\renewcommand{\thetable}{\thesection\arabic{table}}
\renewcommand{\thefigure}{\thesection\arabic{figure}}
\section{Specialists and Evaluation}

\FloatBarrier
\subsection{Specialist Bank and Multi-Domain SFT}\label{sec:specialist-bank}

Base is Qwen3-VL-30B-A3B-Instruct~\citep{Bai2025Qwen3VL}, a
mixture-of-experts vision-language model with 30B total and 3B activated
parameters. It is also the task-vector reference $\theta_{\mathrm{base}}$
of the merge. The STEM, VQA, and embodied specialists are post-trained
from Base with SFT followed by RL, both stages on data from their own
domain. The OCR and GUI
specialists are produced by applying domain RL to two different multi-domain SFT
checkpoints; the OCR specialist's SFT ancestor is the Multi-Domain SFT
model described next (per-domain and per-benchmark scores in
Tables~\ref{tab:main-results} and~\ref{tab:benchmark-scores}).

Multi-Domain SFT is trained from Base in a single supervised run on a
mixture of post-training data from all five domains, without a
reinforcement-learning stage. It is the conventional
single-model alternative to the specialist bank.

\FloatBarrier
\subsection{Evaluation Suite}\label{sec:evaluation-suite}

\begin{table}[H]
\caption{\textsf{\bfseries Evaluation Suite.} The 19 public benchmarks spanning five multimodal domains. Avg.\ assigns equal weight to benchmarks within each domain and then to the five domain means.}\label{tab:evaluation-suite}
\centering\small
\begin{tabularx}{\textwidth}{@{}lX@{}}
\toprule
Domain & Benchmarks \\
\midrule
STEM & We-Math~\citep{Qiao2024WeMath}, MMMU-Pro~\citep{Yue2024MMMUPro}, MathVision~\citep{Wang2024MathVision}, DynaMath~\citep{Zou2024DynaMath} \\
General VQA & MuirBench~\citep{Wang2024MuirBench}, BabyVision~\citep{Chen2026BabyVision}, MMBench English dev v1.1~\citep{Liu2023MMBench}, MMStar~\citep{Chen2024MMStar} \\
Document/OCR/chart & DocVQA validation~\citep{Mathew2020DocVQA}, AI2D~\citep{Kembhavi2016AI2D}, CharXiv reasoning validation~\citep{Wang2024CharXiv}, OCRBench v2~\citep{Fu2025OCRBenchV2} \\
GUI grounding & ScreenSpot-v2~\citep{Wu2024OSAtlas}, ScreenSpot-Pro~\citep{Li2025ScreenSpotPro}, OSWorld-G~\citep{Xie2025OSWorldG} \\
Embodied understanding & MMSI-Bench~\citep{Yang2025MMSI}, RefSpatial-Bench~\citep{Zhou2025RoboRefer}, EmbSpatial-Bench~\citep{Du2024EmbSpatial}, TreeBench~\citep{Wang2025TreeBench} \\
\bottomrule
\end{tabularx}
\end{table}

The suite (Table~\ref{tab:evaluation-suite}, with versions and splits)
combines reasoning, perception, grounding, and spatial tasks so that both
specialist strengths and cross-domain changes are visible. Aggregate
scores over the suite also guide the manual coefficient search for the
tuned merge and set the domain shares for gap-based sampling, whereas the
uniform merge and uniform domain sampling use none of these scores. We
retain benchmark-specific protocols: MMMU-Pro averages its 10-choice and
vision subsets, OSWorld-G measures screenshot grounding rather than
interactive task completion, We-Math uses its strict score, DocVQA uses
ANLS, OCRBench v2 averages its English and Chinese subsets, MMBench
reports single-pass accuracy rather than CircularEval, and DynaMath
reports average rather than worst-case accuracy.

\FloatBarrier
\subsection{Evaluation Protocol}\label{sec:evaluation-protocol}

All models (Base, the specialists, and every student) are evaluated
with the same pipeline. Benchmarks requiring model-based answer scoring
use gpt-oss-20b as a common judge. Generation is greedy, with one
response per input, temperature 0, repetition penalty 1, and a
32,768-token response cap; image preprocessing is identical across
models, and unparseable responses are scored as incorrect.

\textbf{Re-evaluation noise.} We re-evaluated four checkpoints with the
same pipeline: the uniform merge before MOPD and after 50 and 200
updates, and the tuned merge after 50 updates. Across re-evaluations,
Avg.\ changes by at most 0.1 (0.07 RMS), a domain mean by at most 0.7,
and a single benchmark by up to 3.4 points. The noise thresholds of
Section~\ref{sec:setup} follow directly: 0.15 for Avg.\ (about twice the
RMS change) and 0.7 for a domain mean (the largest observed change).
Every reported score is an original evaluation.

\textbf{GUI grounding instruction.} ScreenSpot-v2, ScreenSpot-Pro, and
OSWorld-G share one grounding instruction, applied identically to Base,
the specialists, and every student. Without a system message, each query
is a single user turn containing the instruction below, followed by the
screenshot and the element description (or natural-language instruction
for OSWorld-G) on a new line. The model briefly reasons about the target
location and outputs a JSON click action whose \texttt{x} and \texttt{y}
are integers from 0 to 1,000, normalized over the full screenshot. We fix
this instruction across all models and report it verbatim for
reproducibility. Each paragraph below is passed as a single line
(wrapped here for layout). On ScreenSpot-v2 and ScreenSpot-Pro, the
instruction reads:

\noindent\begin{minipage}{\linewidth}
\begin{quote}\footnotesize
\begin{verbatim}
You are a GUI grounding agent. Given a screenshot and an element
description, locate the element and output a click action.

Briefly reason about the target element location, then output the
click action as JSON.

The expected JSON output format is as follows:

{"action": {"name": "click", "button": "left", "click_type":
"single", "x": 250, "y": 845}}

Here, "name" is the action to perform, "button" is the mouse button,
"click_type" is how the click is performed, and "x" and "y" represent
the target location on screen.

The coordinate system is normalized: output x and y as integers from
0 to 1000, where the entire screenshot maps to this range.
\end{verbatim}
\end{quote}
\end{minipage}

On OSWorld-G, whose queries are natural-language instructions rather
than element descriptions, only the first paragraph differs:

\begin{quote}\footnotesize
\begin{verbatim}
You are a GUI interaction agent. Given a screenshot and a natural
language instruction, identify the target element and output a click
action.
\end{verbatim}
\end{quote}

All models share this instruction, so GUI comparisons are like-for-like;
published Qwen3-VL numbers use a model-specific prompt.

\FloatBarrier
\section{Training Setup}

\FloatBarrier
\subsection{Framework and Main Configuration}
\label{sec:runtime-config}

\begin{figure}[H]\centering
\includegraphics[width=\textwidth]{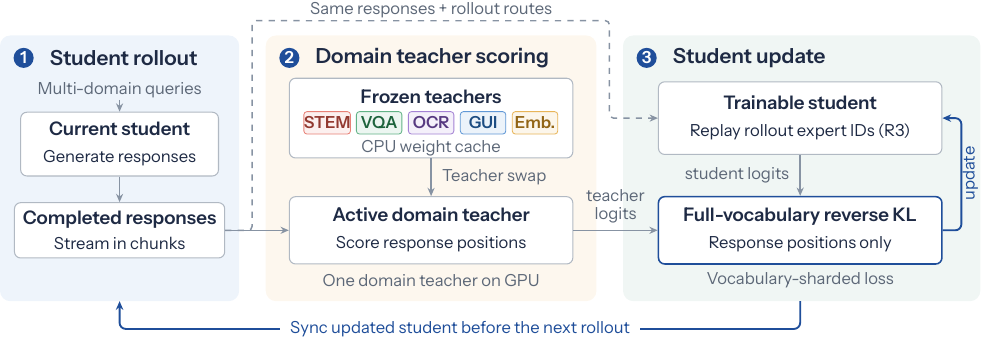}
\caption{\textsf{\bfseries Training Framework.} Rollout generation, teacher scoring, and student updates run as separate components that overlap within each rollout window. Completed responses stream to scoring in chunks, one domain teacher loads from CPU memory onto the training GPUs at a time, and updated weights synchronize before the next rollout. Dashed arrows carry responses and rollout routes; solid arrows carry the training signal.}
\label{fig:training-architecture}
\end{figure}

The training framework (Figure~\ref{fig:training-architecture}) trains the
student with ms-swift~\citep{Zhao2025SWIFT} on the Megatron-LM
backend~\citep{Shoeybi2019MegatronLM} and generates rollouts with a
vLLM engine~\citep{Kwon2023vLLM}. The student generates responses, the
frozen domain teacher of each query scores the same response prefixes,
and the student learns from the teacher's full-vocabulary distribution.
Before each new rollout window, the current student weights are
synchronized to the rollout engine. To isolate single design choices, we
also reimplement the method in verl~\citep{Sheng2025HybridFlow}
(\methodname{}-verl), which trains the student on the Megatron backend of
verl with the same routing replay and supports the supervision sweep and
Open-MOPD comparisons in Appendix~\ref{sec:comparison-controls}.

\textbf{Main experiment.} The main run starts from the uniform merge
and uses uniform domain sampling, AdamW with a constant learning rate of
$3\times10^{-6}$, and a global batch of 512. It runs on three nodes of
eight H200 GPUs each: two nodes train the student, and one serves
rollouts. The objective is full-vocabulary reverse KL averaged over all
valid response tokens in the batch. Excluding initialization and
checkpoint saving, the first 50 updates average 266 seconds each and take
3.7 hours in total. The same run continues to 200 updates.

\textbf{Memory-efficient full-vocabulary distillation.} Teacher and
student materialize vocabulary logits only at response prediction
positions, so the output head is never expanded over the query tokens. The
loss operates on vocabulary shards and processes response rows in chunks,
which shrinks its intermediate tensors while keeping the complete
vocabulary. Frozen teacher weights are cached in CPU memory, and teacher
swapping loads one domain teacher at a time into a reusable model
body on the training GPUs, so the full teacher bank never needs to reside
on the GPUs.

\textbf{Streamed rollouts.} The main experiment streams completed responses to teacher scoring in
chunks while the remaining queries finish. One optimizer update follows
the full rollout window, so a single student policy generates all
responses in that window.

\textbf{Routing consistency.} Teacher scoring reuses the student's
encoded batch, so teacher and student see identical image and token
inputs. R3~\citep{Ma2025R3} replays the rollout expert IDs during student
training while recomputing the gate scores. Together with weight
synchronization before each rollout window, this keeps the training inputs matched to the policy that generated them.

\FloatBarrier
\subsection{Query Construction and Domain Sampling}\label{sec:query-construction}

All policies draw from a common query pool built from the specialists'
post-training data. Each query retains its system and user messages and
any earlier turns, with the final assistant response removed.

Uniform domain sampling assigns a share of 0.2 to each domain. It draws
40,960 queries at random without replacement from each domain to form a
204,800-query pool, and shuffles the pool globally to fix the training
order. Because sampling is random within a domain, each source bucket
contributes in proportion to its size. Dynamic sampling
(Appendix~\ref{sec:runtime-d3}) starts from the same pool.

The gap-based policies use the sampling rule of
Eq.~\eqref{eq:gap-sampling}. On each benchmark of domain $d$, we
subtract the score of the uniform merge from that of the domain teacher
and clip negative differences to zero. The signal $\eta_d$ is the mean of
these clipped gaps across the domain's benchmarks, computed from a
development-stage evaluation and fixed before training, while $L_d$ is
the mean number of valid response tokens per query over the first 50
updates of a uniform-sampling run from the tuned merge. Quotas are
rounded to a total of 25,600 by largest remainder. To separate domain
proportions from sample identity, gap-weighted and length-adjusted gap
sampling are constructed by minimally replacing entries of the uniform
sequence: only the slots needed to meet the new domain quotas are
replaced with unused pool queries from the deficit domains. As a result,
94\% and 90\% of the first 25,600 uniformly sampled queries remain in
place (signals and realized shares in Table~\ref{tab:allocation-shares}).

\begin{table}[H]
\caption{\textsf{\bfseries Sampling Signals and Domain Shares.} Uniform sampling keeps every domain near 20\%, whereas non-uniform policies shift domain shares according to static gaps or online loss. Realized shares are measured over the first 50 updates (25,600 queries); $\eta_d$ is the mean positive teacher--student gap over the uniform merge, and $L_d$ is the mean number of valid response tokens per query.}
\label{tab:allocation-shares}
\centering\small
\setlength{\tabcolsep}{7pt}
\begin{tabular}{@{}lrrrrrr@{}}
\toprule
 & & & \multicolumn{4}{c}{Realized share (\%)} \\
\cmidrule(l){4-7}
Domain & $\eta_d$ & $L_d$ & Uniform & Gap-weighted & Length-adjusted gap & Dynamic \\
\midrule
STEM & 5.0 & 1674.5 & 20.0 & 21.9 & 15.4 & 22.4 \\
VQA & 5.7 & 474.1 & 19.7 & 23.7 & 21.7 & 17.1 \\
OCR & 2.7 & 343.4 & 20.2 & 16.5 & 16.6 & 17.6 \\
GUI & 3.8 & 91.3 & 19.9 & 19.1 & 27.8 & 20.2 \\
Embodied & 3.7 & 368.4 & 20.2 & 18.8 & 18.5 & 22.7 \\
\bottomrule
\end{tabular}
\end{table}

\FloatBarrier
\subsection{Dynamic Domain Sampling}
\label{sec:runtime-d3}

Dynamic sampling follows the D$^3$-MOPD scheduler~\citep{Sun2026D3MOPD},
adapted to our five domains, and multiplies the remaining divergence of
each domain by its recent progress. For domain $d$, let $\kappa_d$ be its
current smoothed KL, $\kappa_{d,0}$ its initial KL, and $\bar\rho_d$ the
mean of its recent signed relative KL changes:
\[
g_d=\frac{\kappa_d}{\kappa_{d,0}},\qquad
v_d=\max(0,-\bar\rho_d).
\]
The scheduler max-normalizes $g_dv_d$ to obtain $u_d$ and instantiates Eq.~\eqref{eq:dynamic-sampling} as
\[
p_d=0.1+0.5\,\operatorname{softmax}_d(u/0.5).
\]
Following the D$^3$-MOPD defaults, the initial KL is averaged over the
first five observations, smoothing uses a 10-observation
exponential-moving-average window, and progress averages up to three
recent relative-change windows. After 20 initial updates of uniform
sampling, the scheduler refreshes every 10 updates to assign new shares
to subsequent batches. Each batch adds a multiplicative jitter
$1+U[-0.3,0.3]$, followed by normalization and largest-remainder integer
rounding.

\FloatBarrier
\subsection{On-Policy Objective Comparison}
\label{sec:runtime-todi}

RKL, FKL, and ToDi use the same uniform-merge initializer, uniform query
sequence, response masking, and token-averaged loss reduction. At a
visited prefix $h$, let $p_v = p_\theta^h(v)$ and $q_v = p_{T_{k(i)}}^h(v)$ denote the student and domain-teacher
probabilities of vocabulary token $v$. FKL uses $\sum_v q_v\log(q_v/p_v)$.
Our implementation of ToDi~\citep{Jung2025ToDi} uses the weight
\[
a_v=\operatorname{stopgrad}
\left[\sigma\!\left(\beta(\log q_v-\log p_v)\right)\right],
\qquad \beta=1,
\]
and its per-prefix loss is
\[
\ell_{\mathrm{ToDi}}=
\sum_v\left[
a_vq_v\log\frac{q_v}{p_v}
+(1-a_v)p_v\log\frac{p_v}{q_v}
\right].
\]
Teacher probabilities and the adaptive weights are detached, so
gradients flow only through the student.

\FloatBarrier
\subsection{Comparison Protocols}\label{sec:comparison-controls}

Query-policy comparisons hold the initializer, teacher bank, batch,
learning rate, RKL objective, and generation limits fixed while changing
query selection, domain sampling, and ordering. Responses are always
generated on-policy.

\textbf{Domain- and source-balanced sampling.} Like uniform domain
sampling, this protocol allocates the same number of queries to each
domain and samples at random, differing only in within-domain selection.
Because the training data of each domain span multiple source buckets,
uniform domain sampling draws from the domain as a whole and reflects
bucket sizes, whereas domain- and source-balanced sampling spreads 5,120
queries per domain evenly across buckets and redistributes any unused
quota from exhausted buckets. All runs under this protocol share training
sources, teacher checkpoints, generation settings, and hardware; it adds
the OCR specialist and an Adam-calibrated tuned merge as starts and
repeats the uniform merge at seeds 42 and 43.

\textbf{Open-MOPD comparison.} All three Open-MOPD runs in
Table~\ref{tab:framework} start from the uniform merge and train with
teacher routing on the same 25,600 uniformly sampled queries (seed 42).
They follow the Open-MOPD recipe~\citep{Gao2026OpenMOPD}: a PPO-clipped
policy-gradient surrogate over the student's top-16 tokens weighted by
student probability, clip ratios 0.2 and 0.28, dual clip 3, no KL
penalty, and token-mean loss reduction within each microbatch. Its three
budget-balancing mechanisms are equal per-domain token-share targets,
multiplicative gap weighting with exponent 1, and reward refresh before
every optimizer update.

As in Open-MOPD, each rollout batch is split into four non-overlapping
minibatches of one update each; using 2,048-query rollout batches gives
512 queries per update as in our method, with a final 1,024-query batch
for two updates so that 50 updates cover all 25,600 queries once. We set
a constant learning rate of $3\times10^{-6}$ rather than pairing
$1.5\times10^{-6}$ with 256-query minibatches so that both recipes train
at the same step size over 50 updates (the default Open-MOPD schedule is
evaluated separately in Table~\ref{tab:openmopd-comparison}). Although
the Open-MOPD paper mentions nucleus truncation ($p=0.99$) on the
student's top-16 tokens, the released code does not apply it under this
token-selection strategy, so none of our runs applies it.

\emph{Released code.} The first run uses the released Open-MOPD
code (commit \texttt{4809a96}) on verl with FSDP2, adapted in four
places; we call it the released code throughout:
\begin{itemize}\setlength{\itemsep}{2pt}\setlength{\parskip}{0pt}
\item \textbf{Multimodal and MoE support:} expert scoring follows the
same sparse path as training, the Qwen3-VL visual position embedding is
sharded with its parent module, and the rollout server is updated for
vLLM~0.11 compatibility.
\item \textbf{Teacher routing:} only the domain teacher scores each
response; the released code scores every response with every teacher and
retains the domain teacher through a one-hot weight, so the training
signal is identical.
\item \textbf{Reward refresh:} the released code deletes the teacher
log-probabilities needed by the refresh before the actor update, which
prevents refresh across multiple updates per rollout batch. We retain
them so that reward refresh runs as published.
\item \textbf{Bookkeeping:} updates are counted and checkpoints saved
per optimizer update, and the final half batch is trained rather than
discarded.
\end{itemize}
Student parameters and optimizer states stay on GPU, with a
one-sequence student microbatch (a larger microbatch would alter how the
token-mean reduction weights sequences). The FSDP2 actor update accounts
for 628 of the 888 seconds per update.

\emph{verl port with routing replay.} Because the released code lacks
rollout routing replay~\citep{Ma2025R3} and verl supports it only on its
Megatron and VeOmni engines, we port the recipe into
\methodname{}-verl (Megatron, tensor parallelism 2, expert parallelism 8),
following the released code and the Open-MOPD equations, and run it with
and without the three budget-balancing mechanisms. Minibatches follow the
released code: each rollout batch is balanced across data-parallel ranks,
and each rank's share is ordered from short to long and back, so the
four updates of a batch see short, long, long, and short responses. The
two backends use 8 and 16 data-parallel ranks, so the exact query
grouping differs.

Three parity checks confirm the port. With one update per batch, the
mechanisms disabled, and the token-averaged loss, its first updates
reproduce the loss of the top-16 run of Table~\ref{tab:supervision}.
With routing replay off, its first two rollout batches match the
released code in loss, per-domain token shares, and gap weights within
run-to-run rollout variation. At update 50, the released code and the
port with routing replay score 68.51 and 68.52, respectively. The port keeps the student on GPU with a one-sequence microbatch and
gives vLLM half of each GPU's memory; it takes 348 seconds per update
with the mechanisms and 351 without them (210 and 211 seconds in the
actor update).

\textbf{Additional Open-MOPD runs.} Across four further FSDP2 runs
without routing replay, \methodname{} outperforms or matches Open-MOPD at
every matched update count, and the strongest Open-MOPD variant requires
120 updates (30,720 queries) to reach the score \methodname{} attains in
50 updates (25,600 queries; Table~\ref{tab:openmopd-comparison}). From
the uniform merge with 512-query rollout batches, reusing each batch for
four full-batch updates reaches 68.7 at update 50 (from 6,656 distinct
queries) and 68.9 at update 200, while making one update per batch on our
query sequence reaches 68.6 at update 50.

From the tuned merge, two runs follow the default Open-MOPD schedule of
1,024 queries split into four 256-query updates at learning rate
$1.5\times10^{-6}$ on four nodes, making two passes over our domain- and
source-balanced queries in 200 updates. Without the three
budget-balancing mechanisms, this run reaches 69.7 only at update 120
(matching the 50-update score of \methodname{}), while the full
Open-MOPD recipe peaks at 69.1.

\textbf{Supervision sweep.} The four \methodname{}-verl runs of
Table~\ref{tab:supervision} start from the uniform merge and share the
25,600 queries in our order, the five domain teachers, one update per
512-query batch, a constant learning rate of $3\times10^{-6}$,
token-averaged loss normalization, and routing replay, differing only in
the per-token objective. Student and teacher log-probabilities are always
normalized over the full vocabulary ($p_\theta$ and $q = p_{T_{k(i)}}$ at
each visited prefix). The sampled-token objective uses the advantage
$A=\mathrm{sg}(\log q(y_t)-\log p_\theta(y_t))$ on the generated token.
The top-$k$ objective selects the student's top $k$ tokens $S$, weights
them by $w=\mathrm{softmax}(\log p_\theta[S])$ (normalized only within
$S$), and applies $A_v=\mathrm{sg}\big(w_v(\log q(v)-\log p_\theta(v))\big)$
with the Open-MOPD dual-clip policy loss (clip 0.2, dual clip 3;
inactive at one update per batch). The full-vocabulary objective is the
exact RKL of Eq.~\eqref{eq:mopd-loss}.

Top-16 and Top-128 both reach Avg.\ 69.6 at update 50 (compared with 68.1
for sampled-token supervision), indicating that increasing the truncation
width adds no further benefit once distribution-level supervision is used.
In these runs, the student trains on two nodes with Megatron
(tensor parallelism 2, expert parallelism 8) and CPU parameter and
optimizer offloading while serving vLLM rollouts on the same two nodes,
and the third node hosts the domain teachers. Per-update time ranges from
412 to 718 seconds (Table~\ref{tab:supervision-timing}), compared with
265.5 seconds on average over the first 50 updates of the main run. In
the main run, generation takes 159.3 seconds, teacher scoring takes
98.4 seconds (including 19.6 seconds to load the domain teacher and
34.5 seconds for its forward pass), the student forward pass takes
32.2 seconds, and weight synchronization takes 12.3 seconds; because
these phases overlap, their sum exceeds the wall-clock total.

\begin{table}[H]
\caption{\textsf{\bfseries Time per Update in the Supervision Sweep.} Our streamed pipeline runs exact full-vocabulary distillation faster than \methodname{}-verl runs any objective, including sampled-token supervision. Shown are mean phase and total times (seconds per update over 50 updates on three nodes, CPU-offloaded student in \methodname{}-verl); Top-$k$ is the extra student forward pass extracting top-$k$ token sets, and Total also includes data movement and weight synchronization. Our pipeline overlaps generation, scoring, and updates (individual phase times in text).}
\label{tab:supervision-timing}
\centering\small
\setlength{\tabcolsep}{7pt}
\begin{tabular}{@{}lrrrrr@{}}
\toprule
Objective & Rollout & Teacher & Top-$k$ & Update & Total \\
\midrule
Sampled token & 75 & 26 & --- & 295 & 412 \\
Student top-16 & 77 & 25 & 91 & 311 & 522 \\
Student top-128 & 78 & 26 & 92 & 312 & 526 \\
Full vocabulary & 77 & 106 & --- & 513 & 718 \\
\midrule
\methodnameaccent{} pipeline & \multicolumn{4}{c}{\textit{overlapped}} & 266 \\
\bottomrule
\end{tabular}
\end{table}

\textbf{Training efficiency.} Because the supervision sweep of
Table~\ref{tab:supervision-timing} offloads student parameters and
optimizer states to CPU, we rerun full-vocabulary \methodname{}-verl for
20 updates with the student kept on GPU and half of each GPU's memory
allocated to vLLM (all other settings unchanged). Over updates 2--15,
that run takes 597.8 seconds per update, compared with 655.2 with CPU
offloading and 263.4 for our framework, giving a 2.3-fold speedup for the
same exact loss. Token throughput is closely matched: over the first 50
updates, our framework processes about 300,000 response tokens per update
and full-vocabulary \methodname{}-verl processes about 324,000.

Relative to the released Open-MOPD code (888 seconds per update), two
design choices account for our 3.3-fold end-to-end speedup across the
complete training stacks: switching to the Megatron backend with routing
replay reduces per-update time from 888 to 348 seconds for the same
Open-MOPD recipe, and our streamed pipeline lowers it further to 266
seconds. All timings follow the matched protocol above, including the
one-sequence student microbatch of the released-code run.

\begin{table}[H]
\caption{\textsf{\bfseries Additional Open-MOPD Runs by Optimizer Update.} \methodname{} matches or outperforms Open-MOPD at every matched update count across both starts. Shown is Avg.\ after the given number of updates (dashes mark unsaved checkpoints across different evaluation grids); none of these FSDP2 Open-MOPD runs uses routing replay. From the tuned merge, Open-MOPD trains with 256 queries per update (two passes over 25,600 queries by update 200).}
\label{tab:openmopd-comparison}
\centering\small
\setlength{\tabcolsep}{3pt}
\begin{tabular*}{\linewidth}{@{\extracolsep{\fill}}lrrrrrrr@{}}
\toprule
Run & 40 & 50 & 80 & 100 & 120 & 150 & 200 \\
\midrule
\multicolumn{8}{@{}l}{\textit{Uniform merge, uniform queries}} \\
\quad Open-MOPD, 4 updates/batch & 68.3 & 68.7 & --- & 68.8 & --- & 68.5 & 68.9 \\
\quad Open-MOPD, 1 update/batch & 68.2 & 68.6 & --- & --- & --- & --- & --- \\
\quad \methodnameaccent{} & 69.0 & \textbf{69.8} & --- & 69.7 & --- & 69.6 & 69.4 \\
\addlinespace[2pt]
\multicolumn{8}{@{}l}{\textit{Tuned merge, domain- and source-balanced queries}} \\
\quad Open-MOPD & 68.6 & --- & 68.9 & --- & 69.1 & --- & 69.0 \\
\quad Open-MOPD w/o mechanisms & 69.1 & --- & 69.6 & --- & \textbf{69.7} & --- & 69.3 \\
\quad \methodnameaccent{} & 69.1 & \textbf{69.7} & --- & --- & --- & --- & --- \\
\bottomrule
\end{tabular*}
\end{table}

\clearpage
\section{Coefficient Calibration}
\label{sec:calibration-details}

\subsection{Protocol and Tuned Merge}
\label{sec:calibration-protocol}

\begin{table}[H]
\caption{\textsf{\bfseries All Merge Starts.} Across all nine evaluated merges, offline calibration does not outperform the uniform merge after 50 MOPD updates. Columns match Table~\ref{tab:merge-variants}; blocks group reference merges, PGD and maximin QP (fit to RKL on teacher trajectories), and the four-objective study (fit to each named objective). Shading and bold follow Table~\ref{tab:main-results}.}
\label{tab:merge-variants-full}
\centering\small
\setlength{\tabcolsep}{9pt}
\begin{tabular}{@{}lrr@{}}
\toprule
Merge & Initial & @50 \\
\midrule
Tuned & \tintcell{avg}{21}{68.5} & \tintcell{avg}{21}{69.8} \\
\midrule
PGD@5 & \tintcell{avg}{3}{65.6} & \tintcell{avg}{3}{68.8} \\
PGD@8 & \tintcell{avg}{6}{66.9} & \tintcell{avg}{12}{69.2} \\
\midrule
QP@2 & \tintcell{avg}{18}{68.2} & \tintcell{avg}{24}{\textbf{69.9}} \\
\midrule
RKL@10 & \tintcell{avg}{6}{66.9} & \tintcell{avg}{9}{69.1} \\
CE@10 & \tintcell{avg}{24}{\textbf{68.7}} & \tintcell{avg}{18}{69.7} \\
FKL@7 & \tintcell{avg}{24}{\textbf{68.7}} & \tintcell{avg}{15}{69.6} \\
ToDi@8 & \tintcell{avg}{12}{67.9} & \tintcell{avg}{6}{68.9} \\
\midrule
Uniform merge & \tintcell{avg}{9}{67.6} & \tintcell{avg}{21}{69.8} \\
\bottomrule
\end{tabular}
\end{table}

Calibration fits on teacher rollouts rather than student rollouts. Using
the ms-swift rollout engine~\citep{Zhao2025SWIFT} from MOPD training,
each teacher generates 1,000 responses on queries from its own domain to
form a 5,000-trajectory pool. The calibration loss averages over response
tokens within each sequence, then over sequences within each domain, and
finally weights the five domains equally. Teachers and DARE masks remain
fixed (drop probability 0.7, mask seed 42), so only the five global
coefficients $\alpha_k$ in Eq.~\eqref{eq:dare-merge} vary, constrained to
$\alpha_k\ge0$ and total strength $\sum_k\alpha_k\in[0.5,2]$.

The uniform merge sets every coefficient to 0.2. Following the manual
search over total scale and DARE drop rate described in
Section~\ref{sec:merge} (Appendix~\ref{sec:coefficient-scale}), the
tuned merge selects the OCR/GUI-up-weighted direction at total scale 1
and drop rate 0.7, the highest-scoring configuration within our manual
sweep on the full 19-benchmark suite before MOPD. Its coefficients are
$(0.10,0.15,0.35,0.30,0.10)$ for the STEM, VQA, OCR, GUI, and embodied
teachers.

From the 5,000-trajectory pool, we select two fitting panels: a
500-trajectory panel (100 per domain) for PGD and common-descent QP, and
a 1,000-trajectory panel (200 per domain) for maximin QP and the
four-objective study. Within the QP family, maximin QP scores higher
before MOPD than common-descent QP (68.2 vs.\ 68.0) and is therefore
continued alongside the other eight merges in
Table~\ref{tab:merge-variants-full}.

\subsection{PGD and QP Optimization}

Let $\mathcal L_k(\alpha)$ be the RKL of domain $k$ on the fitting panel and
$\mathcal L(\alpha)=\frac15\sum_k\mathcal L_k(\alpha)$ its mean.

\textbf{PGD.} Projected gradient descent follows $-\nabla\mathcal L$ and
projects onto the feasible set. The step size is chosen by Armijo
backtracking on $\mathcal L$, starting from an $L_\infty$ step of 0.05 with
sufficient-decrease constant $c=10^{-4}$, shrink factor 0.5, and at most
four trials. The run from the uniform merge accepts eight updates;
PGD@5 and PGD@8 are continued with MOPD.

\textbf{QP.} Both QP variants compute a common descent direction from the
per-domain gradients $g_k=\nabla\mathcal L_k(\alpha)$ by solving
\[
\min_{\delta,\,t}\ \tfrac12\lVert\delta\rVert_2^2-t
\quad\text{s.t.}\quad \tilde g_k^{\top}\delta+t\le0\ \ (k=1,\dots,5)
\]
together with the active feasibility constraints, so that $t$ is the
worst-domain descent margin. Common-descent QP uses normalized gradients
$\tilde g_k=g_k/\lVert g_k\rVert$; maximin QP uses
$\tilde g_k=g_k/\mathcal L_k(\alpha^{\mathrm{eq}})$, which maximizes the
worst relative improvement over the uniform merge $\alpha^{\mathrm{eq}}$.
The direction is rescaled to unit $L_\infty$ norm, and a backtracking
step, starting from an $L_\infty$ length of 0.025, is accepted only if
every domain satisfies the Armijo condition. Maximin QP additionally
requires the fitting loss of every domain to remain at or below that of
the uniform merge, and both variants stop after two accepted updates.

\subsection{Matched Four-Objective Calibration}

All four runs initialize from the uniform merge and perform 10
normalized projected-gradient updates with an $L_\infty$ step of 0.025
on the same 1,000 fitting trajectories without line search, while
tracking RKL on 500 held-out trajectories from the same pool. A shared
RKL-based rule selects the evaluated checkpoint solely from the fitting
and held-out losses (without benchmark scores), choosing update 10 for
RKL and CE, update 7 for FKL, and update 8 for ToDi. Although RKL
calibration reaches the lowest held-out RKL (0.43), its pre-MOPD Avg.\
drops to 66.9; in contrast, CE and FKL both reach 68.7 at held-out RKL
0.64 and 0.68, and ToDi reaches 67.9 at 0.45. At these selected updates,
the five coefficients (STEM, VQA, OCR, GUI, embodied) are
$(0.12,0.05,0.39,0.35,0.41)$ for RKL@10, $(0.15,0.14,0.39,0.39,0.14)$
for CE@10, $(0.18,0.14,0.36,0.34,0.15)$ for FKL@7, and
$(0.14,0.08,0.36,0.34,0.32)$ for ToDi@8, compared with
$(0.10,0.15,0.35,0.30,0.10)$ for the tuned merge.

\subsection{Adam Calibration of the Tuned Merge}

A separate study calibrates the tuned merge with Adam on 500 fitting
trajectories at a learning rate of 0.001 and batch size 50 while tracking
RKL on 100 held-out trajectories (20 per domain). Continuing its
update-8 checkpoint (the Adam-RKL merge) under domain- and
source-balanced sampling (Table~\ref{tab:start-summary}) shows that
calibration raises the initial score of the tuned merge to 68.7, yet
after 50 updates it finishes at 69.3, below the uniform merge (69.5)
under the same protocol.

\subsection{Coefficient Scale}
\label{sec:coefficient-scale}

The uniform merge gives each of the five task vectors a coefficient of
0.2, for a total scale of 1. Keeping the coefficients equal and the DARE
masks fixed, we vary this scale and compare scores before MOPD on a
13-benchmark subset of the evaluation suite that covers all five domains.
Scores rise from 59.5 at scale 0.75 to 61.3 at scale 1 and then level
off (61.7 at scale 1.25), while larger scales are costly: the
merged model generates much longer responses, and the scores fall. At
scale 2, Avg.\ over all 19 benchmarks falls from 67.6 to 61.2, and the
mean response length grows from 787 to 6,587 tokens, with GUI grounding
most affected. At scale 3, several benchmarks collapse, with responses on
MMMU-Pro averaging more than 30,000 tokens.

MOPD restores the overscaled merge. From the scale-3 merge under the
queries, teachers, and optimizer of the main run, mean response length
falls to about 1,000 tokens by update 150, close to the response length
of the main run. Avg.\ reaches 68.9 at update 150 (above Base after 200
updates at 67.7 and above the tuned merge before MOPD at 68.5) and
finishes 0.6 points below the main run at update 200. A scale near 1
therefore reaches the main-run score without the extra updates that an
overscaled merge needs.

The merge is also insensitive to sparsification. At scale 1 on the same
13-benchmark subset, plain task-vector averaging ($\lambda=0$), which
reduces the merge to a uniform model soup of the
specialists~\citep{Wortsman2022Soups,Ilharco2023TaskArithmetic}, scores
61.2. DARE performs closely across all tested drop rates, scoring 61.3,
61.4, 61.3, and 61.7 at $\lambda=0.3$, 0.5, 0.7, and 0.9; our main
comparisons use $\lambda=0.7$.

\FloatBarrier

\clearpage
\section{Additional Results}

\subsection{Benchmark-Level Scores}
\label{sec:teacher-profiles}

\begin{table}[H]
\caption{\textsf{\bfseries Benchmark Scores.} Per-benchmark scores for four starts, five specialists, and \methodname{} (the uniform merge after 50 updates). Shaded cells mark each benchmark's domain teacher; bold marks the row maximum, including ties.}
\label{tab:benchmark-scores}
\centering\footnotesize
\setlength{\tabcolsep}{3pt}
\begin{tabular*}{\textwidth}{@{\extracolsep{\fill}}lrrrrrrrrrr@{}}
\toprule
 & \multicolumn{4}{c}{Starts} & \multicolumn{5}{c}{Specialists} & \\
\cmidrule(lr){2-5}\cmidrule(lr){6-10}
Benchmark & Base & \shortstack{Multi-Domain\\SFT} & Tuned & Uniform & STEM & VQA & OCR & GUI & Emb. & \methodnameaccent{} \\
\midrule
\multicolumn{11}{@{}l}{\textit{STEM}} \\
\quad We-Math & 55.6 & 59.8 & 66.1 & 64.4 & \domaincell{stem}{\textbf{77.0}} & 59.6 & 75.4 & 70.7 & 53.0 & 72.0 \\
\quad MMMU-Pro & 57.0 & 59.9 & 61.8 & 61.8 & \domaincell{stem}{\textbf{64.5}} & 58.0 & 63.7 & 58.8 & 46.0 & 64.1 \\
\quad MathVision & 46.2 & 65.3 & 67.3 & 63.9 & \domaincell{stem}{\textbf{68.4}} & 56.8 & 67.8 & 66.3 & 26.9 & 66.9 \\
\quad DynaMath & 74.4 & 73.9 & 79.5 & 80.0 & \domaincell{stem}{\textbf{81.8}} & 78.1 & 79.1 & 75.0 & 76.7 & 81.1 \\
\multicolumn{11}{@{}l}{\textit{General VQA}} \\
\quad MuirBench & 63.7 & 63.5 & 67.5 & 67.7 & 69.4 & \domaincell{vqa}{\textbf{74.7}} & 65.0 & 66.5 & 64.4 & 73.5 \\
\quad BabyVision & 9.8 & 13.7 & 15.2 & 15.2 & 14.4 & \domaincell{vqa}{\textbf{22.4}} & 13.1 & 14.7 & 11.9 & 17.3 \\
\quad MMBench & 86.9 & 86.5 & 89.2 & 89.4 & 88.2 & \domaincell{vqa}{\textbf{92.0}} & 89.9 & 88.5 & 86.1 & 90.9 \\
\quad MMStar & 72.3 & 75.2 & 76.5 & 76.8 & 76.1 & \domaincell{vqa}{76.3} & 77.3 & 73.9 & 71.1 & \textbf{77.9} \\
\multicolumn{11}{@{}l}{\textit{Document/OCR/chart}} \\
\quad DocVQA & 95.4 & 95.1 & 95.3 & 94.8 & 94.3 & 17.6 & \domaincell{ocr}{\textbf{95.7}} & 95.2 & 91.6 & 93.4 \\
\quad AI2D & 82.6 & 87.3 & 88.4 & 88.3 & 88.2 & 87.4 & \domaincell{ocr}{\textbf{89.8}} & 86.9 & 86.8 & 89.7 \\
\quad CharXiv & 42.6 & 55.9 & 60.6 & 57.2 & 62.3 & 60.5 & \domaincell{ocr}{\textbf{65.8}} & 56.0 & 45.2 & 64.2 \\
\quad OCRBench v2 & 66.6 & 65.8 & 67.4 & \textbf{67.6} & 65.0 & 60.8 & \domaincell{ocr}{67.1} & 67.0 & 64.7 & 63.8 \\
\multicolumn{11}{@{}l}{\textit{GUI grounding}} \\
\quad ScreenSpot-v2 & 86.9 & 95.2 & \textbf{96.1} & 95.8 & 26.2 & 92.8 & 95.4 & \domaincell{gui}{\textbf{96.1}} & 91.8 & \textbf{96.1} \\
\quad ScreenSpot-Pro & 45.4 & 67.1 & 70.9 & 67.1 & 1.5 & 46.9 & 65.6 & \domaincell{gui}{\textbf{71.2}} & 42.4 & 67.3 \\
\quad OSWorld-G & 21.6 & 69.7 & \textbf{71.6} & 68.1 & 20.2 & 41.7 & 67.4 & \domaincell{gui}{\textbf{71.6}} & 45.9 & \textbf{71.6} \\
\multicolumn{11}{@{}l}{\textit{Embodied understanding}} \\
\quad MMSI-Bench & 28.3 & 30.0 & 30.1 & 30.6 & 26.2 & 28.9 & 30.0 & 29.1 & \domaincell{emb}{\textbf{31.5}} & 31.3 \\
\quad RefSpatial & 55.2 & 55.6 & 61.7 & 60.3 & 53.1 & 52.3 & 56.7 & 59.9 & \domaincell{emb}{\textbf{70.4}} & 68.2 \\
\quad EmbSpatial & 77.9 & 79.3 & 80.5 & 80.2 & 79.5 & 79.9 & \textbf{80.6} & 79.5 & \domaincell{emb}{77.9} & 80.4 \\
\quad TreeBench & 42.7 & 41.2 & 45.4 & 46.2 & 47.4 & \textbf{48.6} & 45.9 & 43.7 & \domaincell{emb}{44.2} & 48.4 \\
\midrule
Avg. & 58.1 & 65.9 & 68.5 & 67.6 & 56.0 & 59.8 & 68.4 & 67.5 & 59.4 & \textbf{69.8} \\
\bottomrule
\end{tabular*}
\end{table}

Across the 19 benchmarks in Table~\ref{tab:benchmark-scores}, the narrow
specialists exhibit sharp off-domain drops (for example, the STEM
specialist falls to 1.5 on ScreenSpot-Pro and the VQA specialist to 17.6
on DocVQA). In contrast, \methodname{} eliminates these regressions and
approaches or matches the row maximum across all five domains, with the
largest post-merge gains on We-Math, RefSpatial, and CharXiv.

\subsection{Initialization Trajectories}
\label{sec:trajectory-details}

\begin{figure}[!t]\centering
\includegraphics[width=\textwidth]{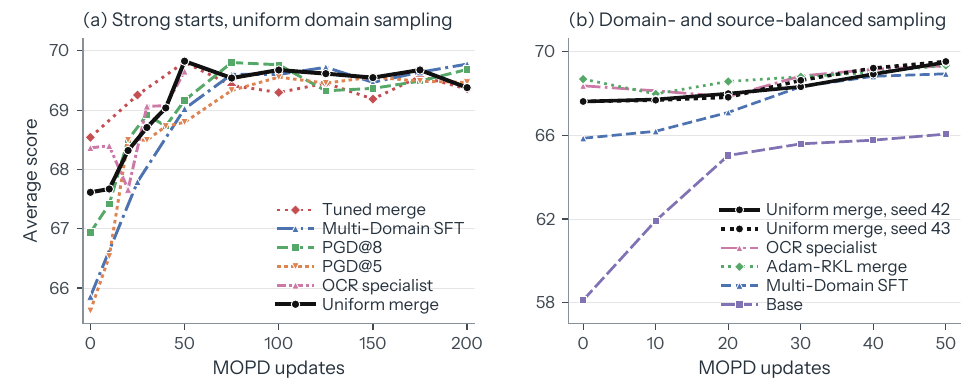}
\caption{\textsf{\bfseries Strong-Start Trajectories.} Strong starts converge under both
training protocols. (a) Uniform domain sampling through 200 updates, including
the PGD-calibrated starts. (b) Domain- and source-balanced sampling
(Appendix~\ref{sec:comparison-controls}) with the uniform merge at
two seeds.}
\label{fig:appendix-trajectories}
\end{figure}

Without coefficient search, the uniform merge matches the tuned merge
after 50 updates (69.8 each), even though the tuned coefficients were
searched on the evaluation suite itself. Under domain- and
source-balanced sampling, it likewise stays within 0.2 points of the
tuned merge (69.5 vs.\ 69.7), and every strong start remains well above
Base (Figure~\ref{fig:appendix-trajectories}). Extending the
domain-sampling comparison of Section~\ref{sec:query-selection} to 200
updates confirms the same pattern (Table~\ref{tab:start-summary}): from
the uniform merge, dynamic and uniform sampling both reach 69.8 at update
50 without further gains by update 200; from Base, gap-weighted and
dynamic sampling reach 67.4 and 67.1 at update 200, against 67.7 under
uniform sampling.

\begin{table}[!t]
\caption{\textsf{\bfseries Summary of All Starts.} Every strong start remains well above Base across all evaluated protocols. Shown is Avg.\ before MOPD (Initial), after 50 updates (with Gain from Initial), and after 200 updates; a dash marks runs trained to 50 updates. Adam-RKL is the update-8 Adam calibration of the tuned merge (balanced-sampling tuned-merge start in Table~\ref{tab:openmopd-comparison}).}
\label{tab:start-summary}
\centering\small
\setlength{\tabcolsep}{9pt}
\begin{tabular}{@{}lrrrr@{}}
\toprule
Start & Initial & @50 & Gain & @200 \\
\midrule
\multicolumn{5}{@{}l}{\textit{Uniform domain sampling}} \\
\quad Base & 58.1 & 66.5 & +8.4 & 67.7 \\
\quad Multi-Domain SFT & 65.9 & 69.0 & +3.2 & 69.8 \\
\quad OCR specialist & 68.4 & 69.7 & +1.3 & --- \\
\quad PGD@5 merge & 65.6 & 68.8 & +3.2 & 69.5 \\
\quad PGD@8 merge & 66.9 & 69.2 & +2.2 & 69.7 \\
\quad Tuned merge & 68.5 & 69.8 & +1.3 & 69.4 \\
\quad Uniform merge & 67.6 & 69.8 & +2.2 & 69.4 \\
\addlinespace[2pt]
\multicolumn{5}{@{}l}{\textit{Gap-weighted domain sampling}} \\
\quad Base & 58.1 & 66.4 & +8.3 & 67.4 \\
\addlinespace[2pt]
\multicolumn{5}{@{}l}{\textit{Dynamic domain sampling}} \\
\quad Base & 58.1 & 66.4 & +8.3 & 67.1 \\
\quad Uniform merge & 67.6 & 69.8 & +2.2 & 69.7 \\
\addlinespace[2pt]
\multicolumn{5}{@{}l}{\textit{Domain- and source-balanced sampling}} \\
\quad Base & 58.1 & 66.1 & +7.9 & --- \\
\quad Multi-Domain SFT & 65.9 & 68.9 & +3.1 & --- \\
\quad OCR specialist & 68.4 & 69.3 & +1.0 & --- \\
\quad Adam-RKL merge & 68.7 & 69.3 & +0.7 & --- \\
\quad Uniform merge, seed 42 & 67.6 & 69.5 & +1.9 & --- \\
\quad Uniform merge, seed 43 & 67.6 & 69.5 & +1.9 & --- \\
\bottomrule
\end{tabular}
\end{table}

\begin{figure}[!t]\centering
\includegraphics[width=\textwidth]{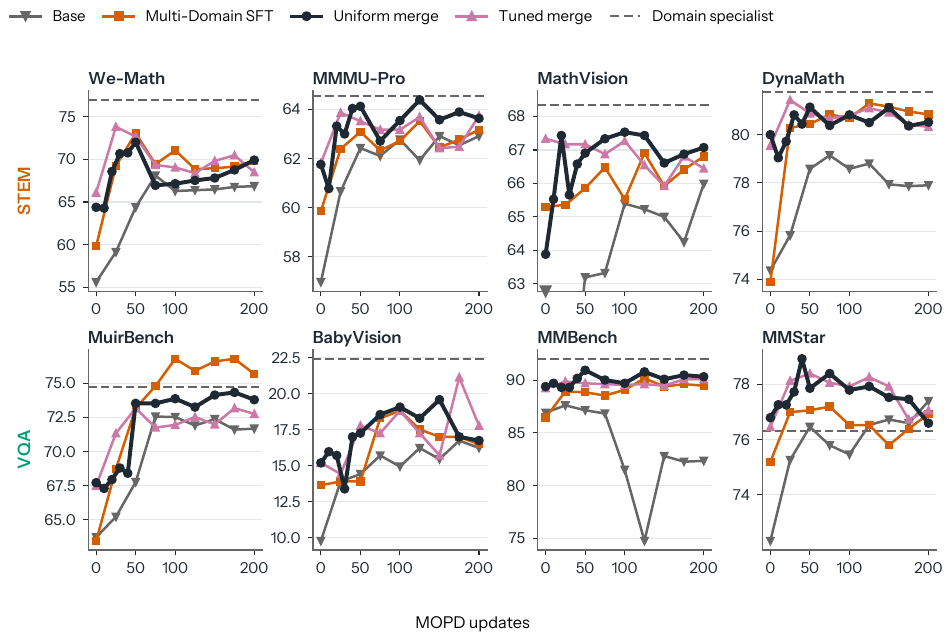}
\caption{\textsf{\bfseries Per-Benchmark Start Trajectories (STEM and VQA).} Strong starts interleave across updates while Base trails on most benchmarks. Shown is the score at every evaluated checkpoint for four starts, with the score of the domain specialist as a dashed reference; a lower-edge triangle marks a Base start below the plotted range.}
\label{fig:start-trajectories-a}
\end{figure}

\begin{figure}[!t]\centering
\includegraphics[width=\textwidth]{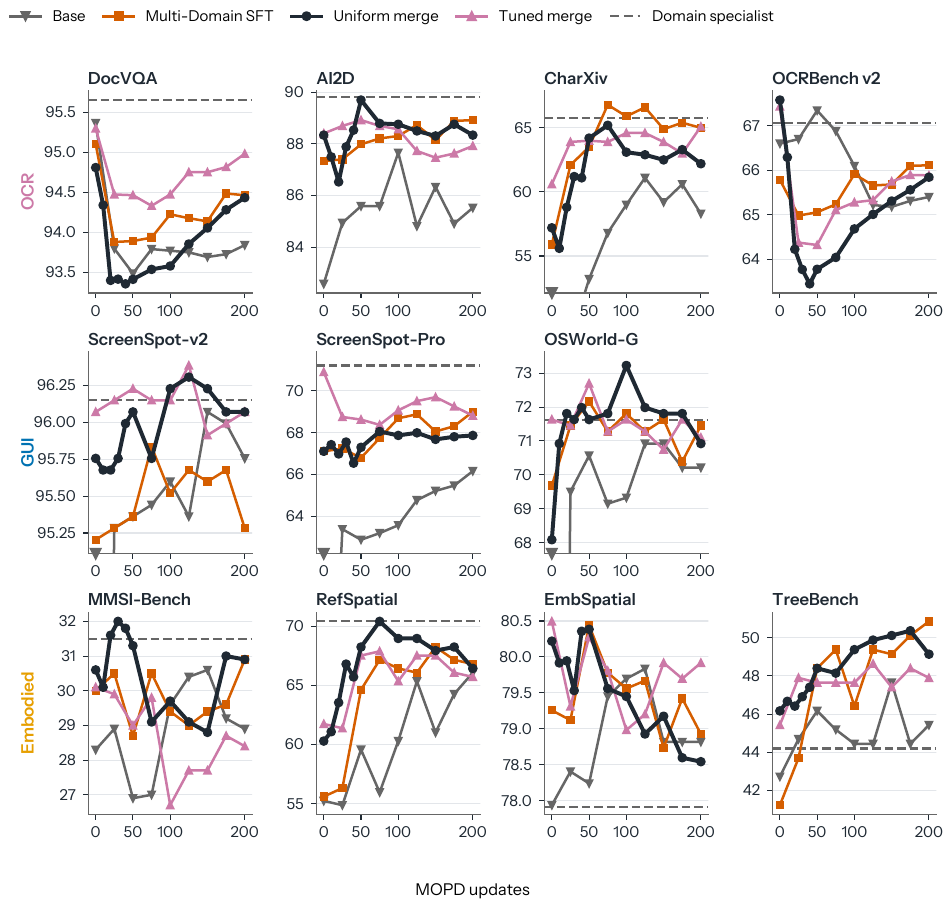}
\caption{\textsf{\bfseries Per-Benchmark Start Trajectories (OCR, GUI, and Embodied).} Same encoding as Figure~\ref{fig:start-trajectories-a}.}
\label{fig:start-trajectories-b}
\end{figure}

At the benchmark level, the strong starts interleave throughout training
(Figures~\ref{fig:start-trajectories-a} and~\ref{fig:start-trajectories-b}).
Averaged over the 19 benchmarks, the spread among the Multi-Domain SFT,
uniform-merge, and tuned-merge starts narrows from 3.0 points before
MOPD to 1.4 after 50 updates and 1.3 after 200. Over the same 50
updates, Base remains below all three strong starts on 15 of the 19
benchmarks. Domain means and Avg., which re-evaluation moves by at most
0.7 and 0.1 (Appendix~\ref{sec:evaluation-protocol}), show the same
convergence.

\begin{figure}[!t]\centering
\includegraphics[width=\textwidth]{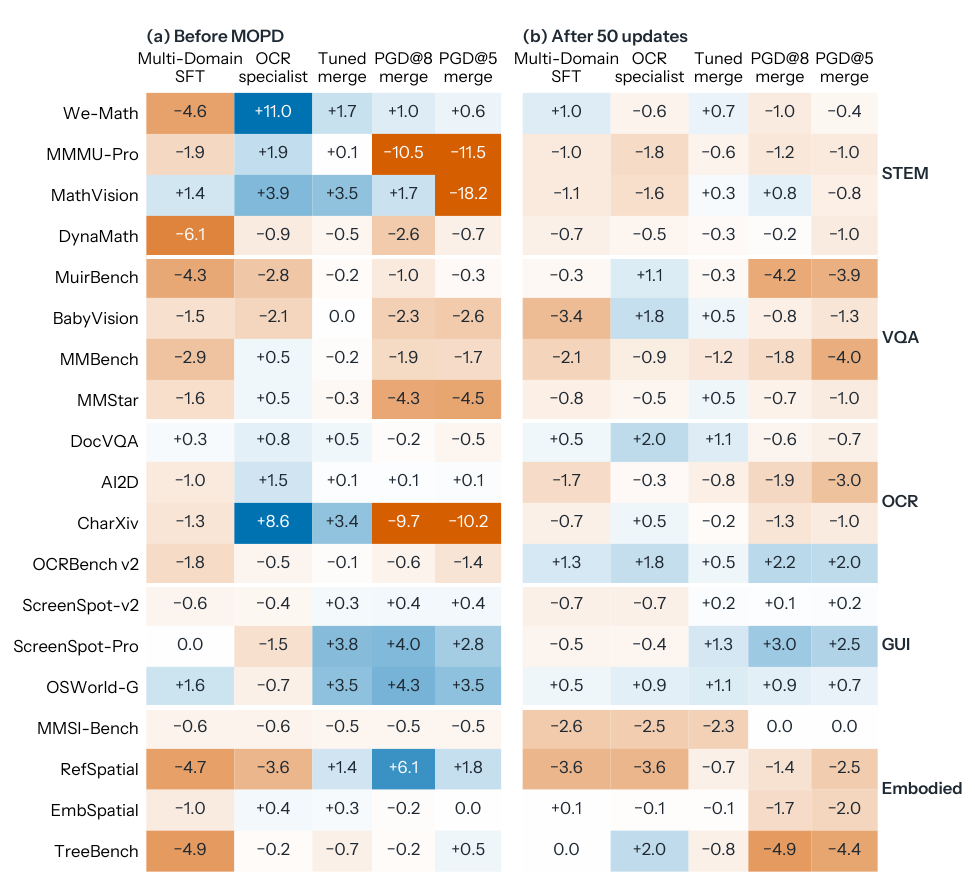}
\caption{\textsf{\bfseries Starts Compared with the Uniform Merge, by Benchmark.} Short MOPD compresses benchmark-level gaps across all alternative starts. Shown is the score difference from the uniform merge for five starts (a) before MOPD and (b) after 50 updates (blue: higher than the uniform merge; vermilion: lower).}
\label{fig:start-differences}
\end{figure}

Short MOPD also narrows benchmark-level differences between the uniform
merge and the five alternative starts (Figure~\ref{fig:start-differences}).
Before MOPD, 64 of the 95 differences are within 2 points, and the
largest reaches 18.2 points (MathVision under the PGD@5 merge). After 50
updates, 75 of the 95 lie within 2 points and the largest drops to 4.9.

\subsection{Domain Trajectory through 200 Updates}

In the main run from the uniform merge, 50 updates suffice even though
the training loss keeps falling (Table~\ref{tab:main-run-trajectory}).
From update 75 through update 200, the average score remains between 69.4
and 69.7 while the training RKL (averaged over the 10 updates ending at
each checkpoint) drops from 0.080 to 0.049 with no further capability
gain.

\begin{table}[!htbp]
\caption{\textsf{\bfseries Main-Run Trajectory.} Average score saturates by update 50 even as training RKL falls from 0.105 to 0.049. Shown are domain means, Avg., and training RKL (averaged over the preceding 10 updates) from update 0 (uniform merge) through update 200, with the best-specialist reference at the bottom.}
\label{tab:main-run-trajectory}
\centering\small
\setlength{\tabcolsep}{7pt}
\begin{tabular}{@{}lrrrrrrr@{}}
\toprule
Update & STEM & VQA & OCR & GUI & Emb. & Avg. & RKL \\
\midrule
0 & 67.5 & 62.3 & 77.0 & 77.0 & 54.3 & 67.6 & --- \\
10 & 67.4 & 62.6 & 75.9 & 78.0 & 54.4 & 67.7 & 0.318 \\
20 & 69.8 & 62.6 & 75.7 & 78.2 & 55.4 & 68.3 & 0.208 \\
30 & 70.0 & 62.3 & 76.6 & 78.3 & 56.3 & 68.7 & 0.145 \\
40 & 70.5 & 63.6 & 76.6 & 78.2 & 56.3 & 69.0 & 0.121 \\
50 & 71.0 & 64.9 & 77.8 & 78.3 & 57.1 & 69.8 & 0.105 \\
75 & 69.3 & 65.1 & 77.9 & 78.5 & 56.8 & 69.5 & 0.080 \\
100 & 69.8 & 65.1 & 77.5 & 79.1 & 56.9 & 69.7 & 0.065 \\
125 & 70.0 & 65.1 & 77.6 & 78.8 & 56.7 & 69.6 & 0.056 \\
150 & 69.8 & 65.3 & 77.5 & 78.6 & 56.5 & 69.5 & 0.055 \\
175 & 70.0 & 64.8 & 78.0 & 78.6 & 57.0 & 69.7 & 0.054 \\
200 & 70.3 & 64.4 & 77.7 & 78.3 & 56.3 & 69.4 & 0.049 \\
\midrule
Best-specialist ref. & 72.9 & 66.4 & 79.6 & 79.7 & 56.0 & 70.9 & --- \\
\bottomrule
\end{tabular}
\end{table}

\begin{figure}[!t]\centering
\includegraphics[width=\textwidth]{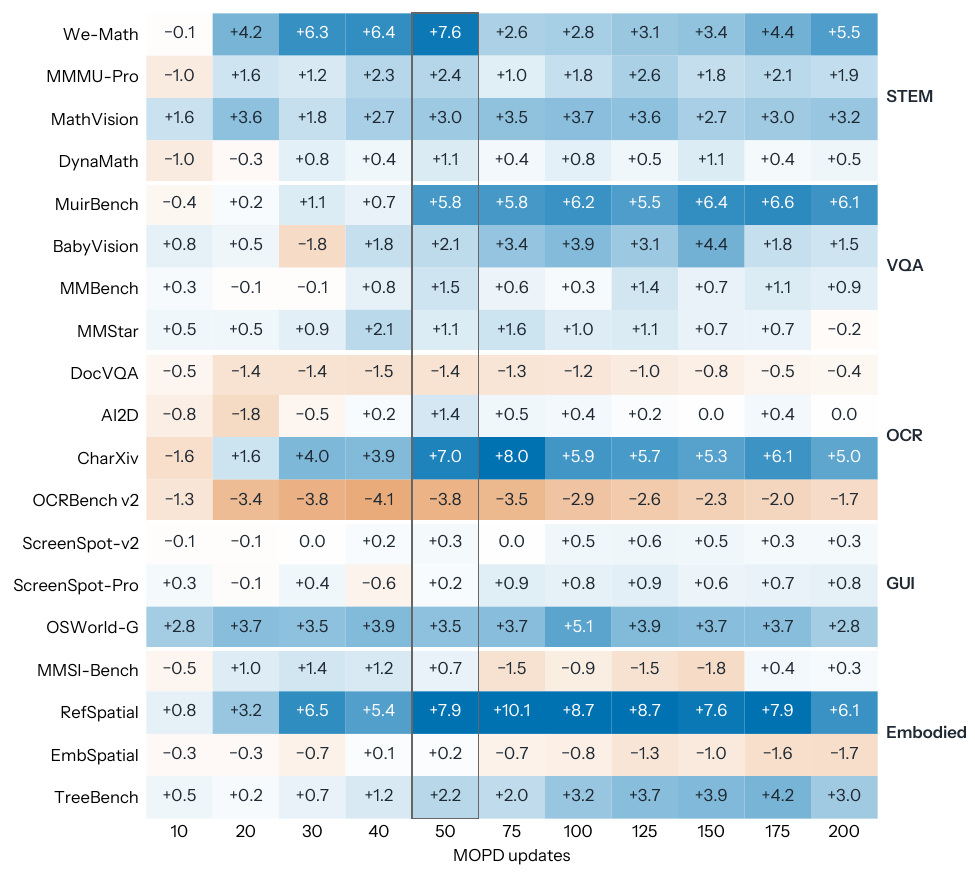}
\caption{\textsf{\bfseries Benchmark Gains by Update.} Major benchmark gains form largely by update 50 and remain stable thereafter. Shown is the score change from the uniform merge at every evaluated checkpoint of the main run; the outline marks update 50.}
\label{fig:gain-timing}
\end{figure}

At the benchmark level, gains concentrate on a few tasks and largely
form by update 50 (Figure~\ref{fig:gain-timing}). By update 50, We-Math
gains 7.6 points, RefSpatial gains 7.9 (peaking at 10.1 at update 75),
CharXiv gains 7.0 (peaking at 8.0 at update 75), and MuirBench gains
5.8, and the run holds those levels through update 200. Meanwhile, ScreenSpot-v2,
DynaMath, and MMBench stay within about 1.5 points throughout, whereas
DocVQA and OCRBench v2 dip early and recover partly by update 200
($-1.4$ to $-0.4$ and $-3.8$ to $-1.7$).

\subsection{Teacher-Generated Responses}\label{sec:offpolicy}

To isolate the contribution of on-policy responses, we compare the main
run with an off-policy variant trained on fixed responses generated once
per query by each domain teacher prior to training. The variant shares
the uniform-merge start, the 25,600 queries in the same order, teacher
routing, full-vocabulary RKL with identical normalization, and the same
optimizer settings; without student rollouts, neither a rollout engine
nor routing replay applies. Training on on-policy student responses
raises the 50-update Avg.\ from 68.6 to 69.8 (Table~\ref{tab:offpolicy}),
a gain of 1.2 points, outperforming the off-policy run in four domains
and matching it in GUI (78.3 vs.\ 78.5, within domain noise).

\begin{table}[!htbp]
\caption{\textsf{\bfseries Student vs.\ Teacher Responses.} On-policy student responses outperform fixed teacher responses by 1.2 points after 50 updates. Both runs start from the uniform merge (Avg.\ 67.6) with identical queries, teachers, and loss; the off-policy variant omits student rollouts and routing replay. Columns report Avg.\ across updates and domain means at update 50.}
\label{tab:offpolicy}
\centering\small
\setlength{\tabcolsep}{5pt}
\begin{tabular}{@{}lrrrrrrrrr@{}}
\toprule
 & \multicolumn{4}{c}{Avg.\ after updates} & \multicolumn{5}{c}{Domain means at 50} \\
\cmidrule(lr){2-5}\cmidrule(l){6-10}
Responses & 20 & 30 & 40 & 50 & STEM & VQA & OCR & GUI & Emb. \\
\midrule
Teacher & 67.6 & 67.9 & 68.2 & 68.6 & 68.7 & 64.1 & 76.4 & 78.5 & 55.5 \\
Student (\methodnameaccent{}) & 68.3 & 68.7 & 69.0 & \textbf{69.8} & 71.0 & 64.9 & 77.8 & 78.3 & 57.1 \\
\bottomrule
\end{tabular}
\end{table}

\subsection{Open-MOPD by Domain and Benchmark}\label{sec:openmopd-detail}

\begin{figure}[!t]\centering
\includegraphics[width=\textwidth]{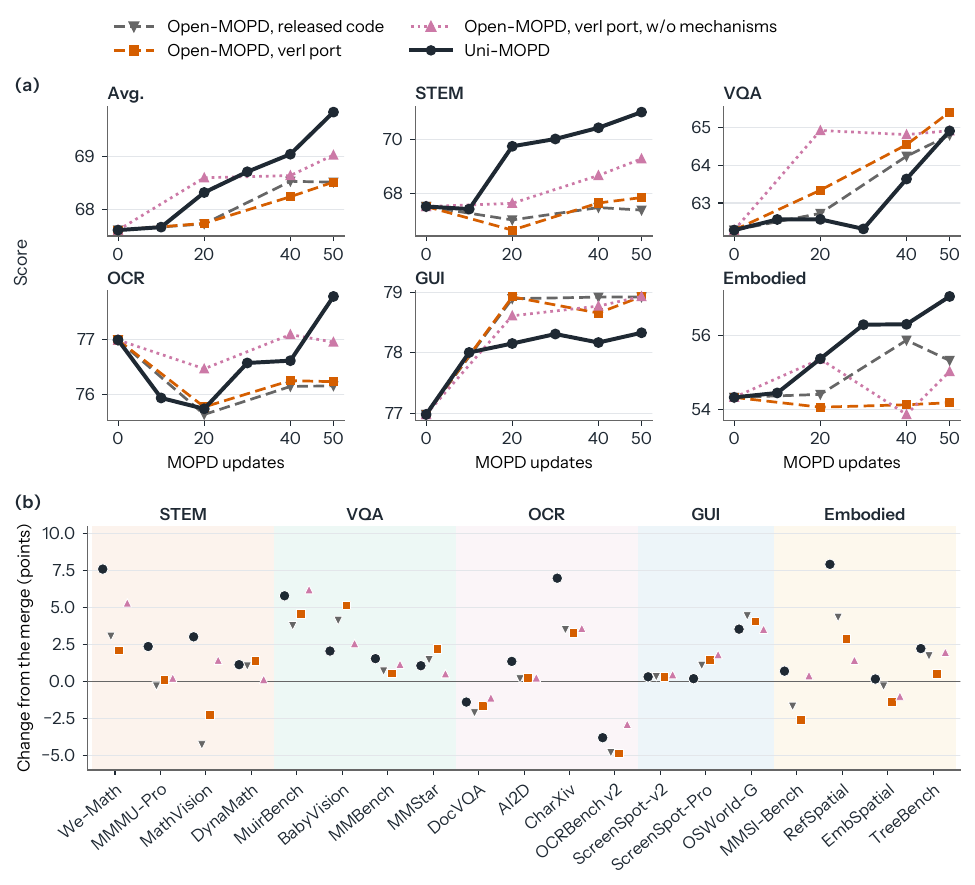}
\caption{\textsf{\bfseries Open-MOPD by Domain and Benchmark.} \methodname{} outperforms Open-MOPD most strongly in STEM, OCR, and Embodied. (a) Avg.\ and domain means over the first 50 updates from the uniform merge on the same queries. (b) Score change from the uniform merge after 50 updates on each benchmark.}
\label{fig:openmopd-detail}
\end{figure}

\methodname{} outperforms Open-MOPD most strongly in STEM, with clear
gains in OCR and Embodied as well. Because all runs start from the
uniform merge and train on the same queries, the curves of
Figure~\ref{fig:openmopd-detail}a differ only in the recipe. In STEM,
\methodname{} reaches 71.0 at update 50, against 67.4 for the released
code and 67.8 for the verl port, both near the initial score of the
uniform merge (67.5); removing the three budget-balancing mechanisms
lifts the verl port to 69.3. Embodied follows the same ordering (57.1
against 55.3 and 54.2), and OCR is 1.6 points higher than under both
Open-MOPD runs (77.8 against 76.2), while VQA and GUI remain comparable
within the 0.7-point domain noise band. At the benchmark level
(Figure~\ref{fig:openmopd-detail}b), the STEM and Embodied gaps come
from a few tasks: MathVision changes by $+3.0$ under \methodname{}
against $-4.3$ and $-2.3$ under the two Open-MOPD runs. Similarly,
We-Math gains $+7.6$ against $+3.0$ and $+2.1$, and RefSpatial gains
$+7.9$ against $+4.3$ and $+2.9$.

\FloatBarrier
\subsection{A Second Specialist Bank}\label{sec:four-teacher}

To verify that the merge finding does not rely on the main bank's two
broad specialists (OCR and GUI, whose RL starts from multi-domain SFT
checkpoints; Section~\ref{sec:specialists}), we evaluate a second bank
whose four specialists are trained strictly with SFT and RL on their own
domains. It keeps the STEM, VQA, and embodied specialists, replaces the
OCR specialist with one trained by OCR-domain SFT and RL, and excludes
GUI, using the 16 benchmarks of the four remaining domains with Avg.\
defined as the mean of their four domain means
(Table~\ref{tab:bank2-teachers}). Across four starts (Base, the
strongest single specialist STEM at Avg.\ 66.0, and uniform merges at
scales 1 and 1.25), all runs share 102,400 uniformly sampled queries in
the same order and the main MOPD recipe for 200 updates, so only the
start differs.

\begin{table}[H]
\caption{\textsf{\bfseries Specialists in the Second Bank.} Domain means and Avg.\ across 16 benchmarks for four narrow specialists trained strictly on their own domains (Table~\ref{tab:main-results} STEM, VQA, and embodied specialists plus an OCR-only specialist). Bold marks each specialist's own-domain score; shading ranks specialists within each column; the best-specialist reference averages the four bold entries.}
\label{tab:bank2-teachers}
\centering\small
\setlength{\tabcolsep}{7pt}
\begin{tabular}{@{}lrrrrr@{}}
\toprule
Specialist & STEM & VQA & OCR & Emb. & Avg. \\
\midrule
STEM & \tintcell{stem}{24}{\textbf{72.9}} & \tintcell{vqa}{17}{62.0} & \tintcell{ocr}{17}{77.5} & \tintcell{emb}{10}{51.5} & \tintcell{avg}{24}{66.0} \\
VQA & \tintcell{stem}{17}{63.1} & \tintcell{vqa}{24}{\textbf{66.4}} & \tintcell{ocr}{3}{56.6} & \tintcell{emb}{17}{52.4} & \tintcell{avg}{10}{59.6} \\
OCR & \tintcell{stem}{10}{60.5} & \tintcell{vqa}{10}{59.3} & \tintcell{ocr}{24}{\textbf{78.3}} & \tintcell{emb}{3}{50.3} & \tintcell{avg}{17}{62.1} \\
Embodied & \tintcell{stem}{3}{50.6} & \tintcell{vqa}{3}{58.4} & \tintcell{ocr}{10}{72.1} & \tintcell{emb}{24}{\textbf{56.0}} & \tintcell{avg}{3}{59.3} \\
\midrule
Best-specialist ref. & 72.9 & 66.4 & 78.3 & 56.0 & 68.4 \\
\bottomrule
\end{tabular}
\end{table}

\begin{table}[H]
\caption{\textsf{\bfseries Second Specialist Bank.} Uniform merges outperform Base and reach the shared performance level in half the updates of the strongest single specialist. Shown is Avg.\ over 16 benchmarks across MOPD updates under identical queries and recipe; shading and bold follow Table~\ref{tab:main-results}.}
\label{tab:four-teacher}
\centering\small
\setlength{\tabcolsep}{7pt}
\begin{tabular}{@{}lrrrrrr@{}}
\toprule
 & \multicolumn{6}{c}{Avg.\ after updates} \\
\cmidrule(l){2-7}
Start & 0 & 25 & 50 & 100 & 150 & 200 \\
\midrule
Base & \tintcell{avg}{3}{59.8} & \tintcell{avg}{3}{62.5} & \tintcell{avg}{3}{63.8} & \tintcell{avg}{3}{65.5} & \tintcell{avg}{10}{65.6} & \tintcell{avg}{3}{65.0} \\
STEM specialist & \tintcell{avg}{24}{\textbf{66.0}} & \tintcell{avg}{24}{\textbf{65.7}} & \tintcell{avg}{14}{66.3} & \tintcell{avg}{10}{66.2} & \tintcell{avg}{3}{65.2} & \tintcell{avg}{24}{\textbf{67.9}} \\
Uniform merge, scale 1.25 & \tintcell{avg}{17}{65.4} & \tintcell{avg}{17}{63.0} & \tintcell{avg}{24}{\textbf{66.6}} & \tintcell{avg}{24}{\textbf{67.7}} & \tintcell{avg}{24}{\textbf{67.2}} & \tintcell{avg}{14}{66.9} \\
Uniform merge (\methodnameaccent{}) & \tintcell{avg}{10}{65.0} & \tintcell{avg}{10}{62.6} & \tintcell{avg}{24}{\textbf{66.6}} & \tintcell{avg}{17}{67.6} & \tintcell{avg}{17}{66.6} & \tintcell{avg}{14}{66.9} \\
\midrule
Best-specialist ref. & \multicolumn{6}{c}{68.4} \\
\bottomrule
\end{tabular}
\end{table}

\begin{table}[H]
\caption{\textsf{\bfseries Domain Means in the Second Bank.} Domain means before MOPD (Initial) and after 50 updates for the four starts of Table~\ref{tab:four-teacher}, with the best specialist per domain for reference. Shading ranks starts within each column and block; bold marks the highest 50-update value.}
\label{tab:four-teacher-domains}
\centering\small
\setlength{\tabcolsep}{7pt}
\begin{tabular}{@{}lrrrrr@{}}
\toprule
Start & STEM & VQA & OCR & Emb. & Avg. \\
\midrule
\multicolumn{6}{@{}l}{\textit{Initial}} \\
Base & \tintcell{stem}{3}{58.3} & \tintcell{vqa}{3}{58.2} & \tintcell{ocr}{3}{71.8} & \tintcell{emb}{3}{51.0} & \tintcell{avg}{3}{59.8} \\
STEM specialist & \tintcell{stem}{24}{72.9} & \tintcell{vqa}{10}{62.0} & \tintcell{ocr}{17}{77.5} & \tintcell{emb}{10}{51.5} & \tintcell{avg}{24}{66.0} \\
Uniform merge, scale 1.25 & \tintcell{stem}{17}{67.1} & \tintcell{vqa}{17}{62.4} & \tintcell{ocr}{24}{77.5} & \tintcell{emb}{24}{54.6} & \tintcell{avg}{17}{65.4} \\
Uniform merge (\methodnameaccent{}) & \tintcell{stem}{10}{66.4} & \tintcell{vqa}{24}{62.4} & \tintcell{ocr}{10}{77.1} & \tintcell{emb}{17}{54.0} & \tintcell{avg}{10}{65.0} \\
\addlinespace[2pt]
\multicolumn{6}{@{}l}{\textit{After 50 updates}} \\
Base & \tintcell{stem}{3}{68.2} & \tintcell{vqa}{3}{61.4} & \tintcell{ocr}{3}{75.3} & \tintcell{emb}{3}{50.4} & \tintcell{avg}{3}{63.8} \\
STEM specialist & \tintcell{stem}{24}{\textbf{71.3}} & \tintcell{vqa}{17}{64.4} & \tintcell{ocr}{24}{\textbf{77.6}} & \tintcell{emb}{10}{51.8} & \tintcell{avg}{10}{66.3} \\
Uniform merge, scale 1.25 & \tintcell{stem}{10}{69.5} & \tintcell{vqa}{24}{\textbf{64.8}} & \tintcell{ocr}{10}{76.9} & \tintcell{emb}{24}{\textbf{55.2}} & \tintcell{avg}{24}{\textbf{66.6}} \\
Uniform merge (\methodnameaccent{}) & \tintcell{stem}{17}{70.1} & \tintcell{vqa}{10}{64.0} & \tintcell{ocr}{17}{77.2} & \tintcell{emb}{17}{55.0} & \tintcell{avg}{24}{\textbf{66.6}} \\
\midrule
Best-specialist ref. & 72.9 & 66.4 & 78.3 & 56.0 & 68.4 \\
\bottomrule
\end{tabular}
\end{table}

\begin{table}[H]
\caption{\textsf{\bfseries Generation Truncation in the Second Bank.} Both merge starts stabilize response length quickly, falling to around 1\% truncation by update 100 after a transient peak at update 25. Shown is the percentage of AI2D answers hitting the 32,768-token generation cap; the truncation rate of Base keeps rising, and the STEM-specialist start peaks at update 150.}
\label{tab:bank2-truncation}
\centering\small
\setlength{\tabcolsep}{7pt}
\begin{tabular}{@{}lrrrrr@{}}
\toprule
Start & 25 & 50 & 100 & 150 & 200 \\
\midrule
Uniform merge (\methodnameaccent{}) & 23.4 & 4.3 & 1.0 & 0.2 & 0.4 \\
Uniform merge, scale 1.25 & 23.3 & 5.0 & 1.1 & 0.9 & 0.3 \\
STEM specialist & 2.6 & 6.8 & 41.1 & 75.9 & 0.0 \\
Base & 6.6 & 10.5 & 40.9 & 52.2 & 52.9 \\
\bottomrule
\end{tabular}
\end{table}

The starting-point and short-budget findings extend to this bank
(Table~\ref{tab:four-teacher}). Without training, the merge (65.0)
already matches the score that Base attains after 200 updates (65.0) and
remains 1.1 points above the update-50 score of Base. After 50 updates,
both merges outperform Base by 2.7--2.8 points and the STEM-specialist
start by 0.3 points. As in the main bank, the strong starts converge to
a similar level, but the merges arrive much sooner: both attain
67.6--67.7 by update 100, outperforming the STEM-specialist start by
1.4--1.5 points at that checkpoint. The STEM start matches this level
only at update 200 (67.9), after its AI2D truncation rate climbs to 76\%
(Table~\ref{tab:bank2-truncation}). The merges thus achieve the shared
performance level in half as many updates, and scales 1 and 1.25 follow
nearly identical trajectories. In this narrow bank, the embodied gain
comes from the embodied teacher alone
(Table~\ref{tab:four-teacher-domains}), whereas in the main bank other
specialists lead on part of this domain and lift the embodied mean above
its teacher (Section~\ref{sec:what-mopd-adds}).

The merges also stabilize long-response generation quickly. On AI2D,
their truncation rate at the 32,768-token generation cap peaks near 23\%
at update 25, falls to about 5\% at update 50, and remains at or below
1\% from update 100 onward (Table~\ref{tab:bank2-truncation}). In
contrast, the truncation rate of the STEM-specialist start rises to 41\%
and 76\% at updates 100 and 150 before returning to zero at update 200,
while Base remains at 41--53\% from update 100 onward.

\FloatBarrier

\end{document}